\documentclass[10pt,twocolumn,letterpaper]{article}

\usepackage{cvpr}
\usepackage{times}
\usepackage{epsfig}
\usepackage{graphicx}
\usepackage{amsmath}
\usepackage{amssymb}

\def\swthree{0.33\linewidth}
\def\swfour{0.25\linewidth}

\def\swseven{0.14\linewidth}

\usepackage[pagebackref=true,breaklinks=true,letterpaper=true,colorlinks,bookmarks=false]{hyperref}

\cvprfinalcopy 

\def\cvprPaperID{1804} 

\ifcvprfinal\fi
\begin{document}
    
    
        \title{Learning a Reinforced Agent for Flexible Exposure Bracketing Selection}
        
        \author{ Zhouxia Wang \textsuperscript{1,2}, Jiawei Zhang\textsuperscript{1}\thanks{Corresponding author is Jiawei Zhang (Email: zhjw1988@gmail.com). This work was partially supported by HKU Seed Fund for Basic Research, Start-up Fund and Research Donation from SenseTime.}, Mude Lin\textsuperscript{1},Jiong Wang\textsuperscript{1},Ping Luo\textsuperscript{2},Jimmy Ren\textsuperscript{1} \\
        \and \textsuperscript{1}SenseTime Research, \textsuperscript{2}The University of Hong Kong \\
                {\tt\scriptsize \{zhouzi1212, zhjw1988, jimmy.sj.ren\}@gmail.com, \{linmude, wangjiong\}@sensetime.com, pluo@cs.hku.hk }}

    \maketitle
    \thispagestyle{empty}
    
        \begin{abstract}
            Automatically selecting exposure bracketing (images exposed differently) is important to obtain a high dynamic range image by using multi-exposure fusion.
            Unlike previous methods that have many restrictions such as requiring camera response function, sensor noise model, and a stream of preview images with different exposures (not accessible in some scenarios \eg some mobile applications), we propose a novel deep neural network to automatically select exposure bracketing, named EBSNet, which is sufficiently flexible without having the above restrictions.
            EBSNet is formulated as a reinforced agent that is trained by maximizing rewards provided by a multi-exposure fusion network (MEFNet).
            %
            By utilizing the illumination and semantic information extracted from just a single auto-exposure preview image, EBSNet can select an optimal exposure bracketing for multi-exposure fusion.
            EBSNet and MEFNet can be jointly trained to produce favorable results against recent state-of-the-art approaches.
            To facilitate future research, we provide a new benchmark dataset for multi-exposure selection and fusion.
            Our code and proposed benchmark dataset will be released in https://github.com/wzhouxiff/EBSNetMEFNet.git
        \end{abstract}
        
            \section{Introduction}
            
            Many real-world scenes have a very large dynamic range.
            However, the digital cameras, especially those on mobile devices, have limited dynamic range making them infeasible to capture a wide dynamic range image with only a single shot.
            Stack-based high dynamic range (HDR) imaging is a widely used technique to solve this issue by fusing several low dynamic range (LDR) images captured at different exposures~\cite{cai2018learning, debevec2008recovering, kou2017multi, li2013image, ma2017robust, mertens2009exposure, prabhakar2017deepfuse, raman2009bilateral, shen2011generalized}.
            For example, Debevec et al.~\cite{debevec2008recovering} propose to use geometrically different exposure settings.
            However, this approach leads to a large exposure bracketing that has both long acquisition and post-processing time, high memory requirement, as well as ghost artifacts yielded by motion.
            Therefore, selecting a small exposure bracketing is a key and important problem.
            
            \begin{figure}[t]\footnotesize
            \begin{center}
            \renewcommand{\tabcolsep}{1pt}
            \begin{tabular}{ccc}
                \includegraphics[width=\swthree,angle=0]{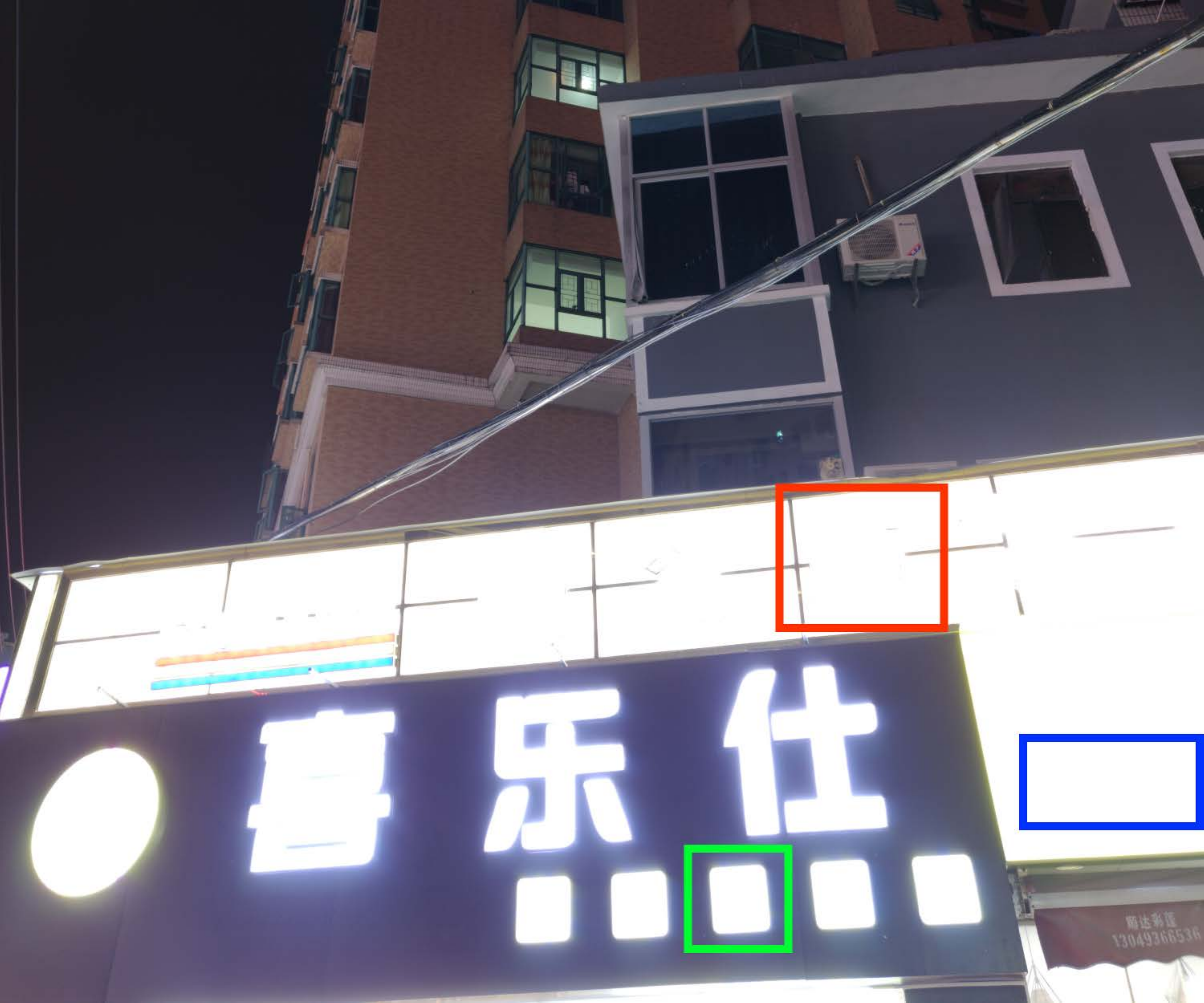} &
                \includegraphics[width=\swthree,angle=0]{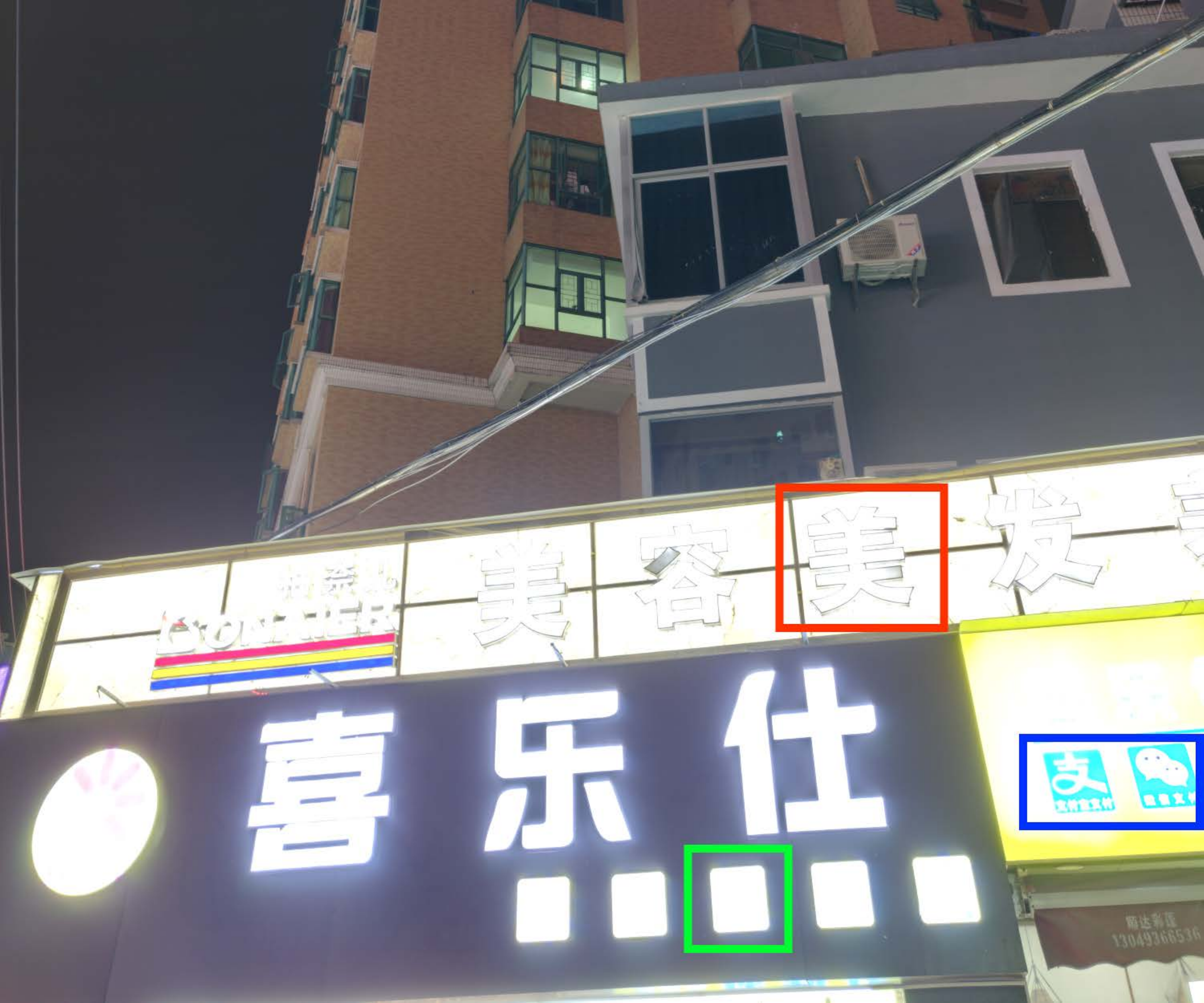} &
                \includegraphics[width=\swthree,angle=0]{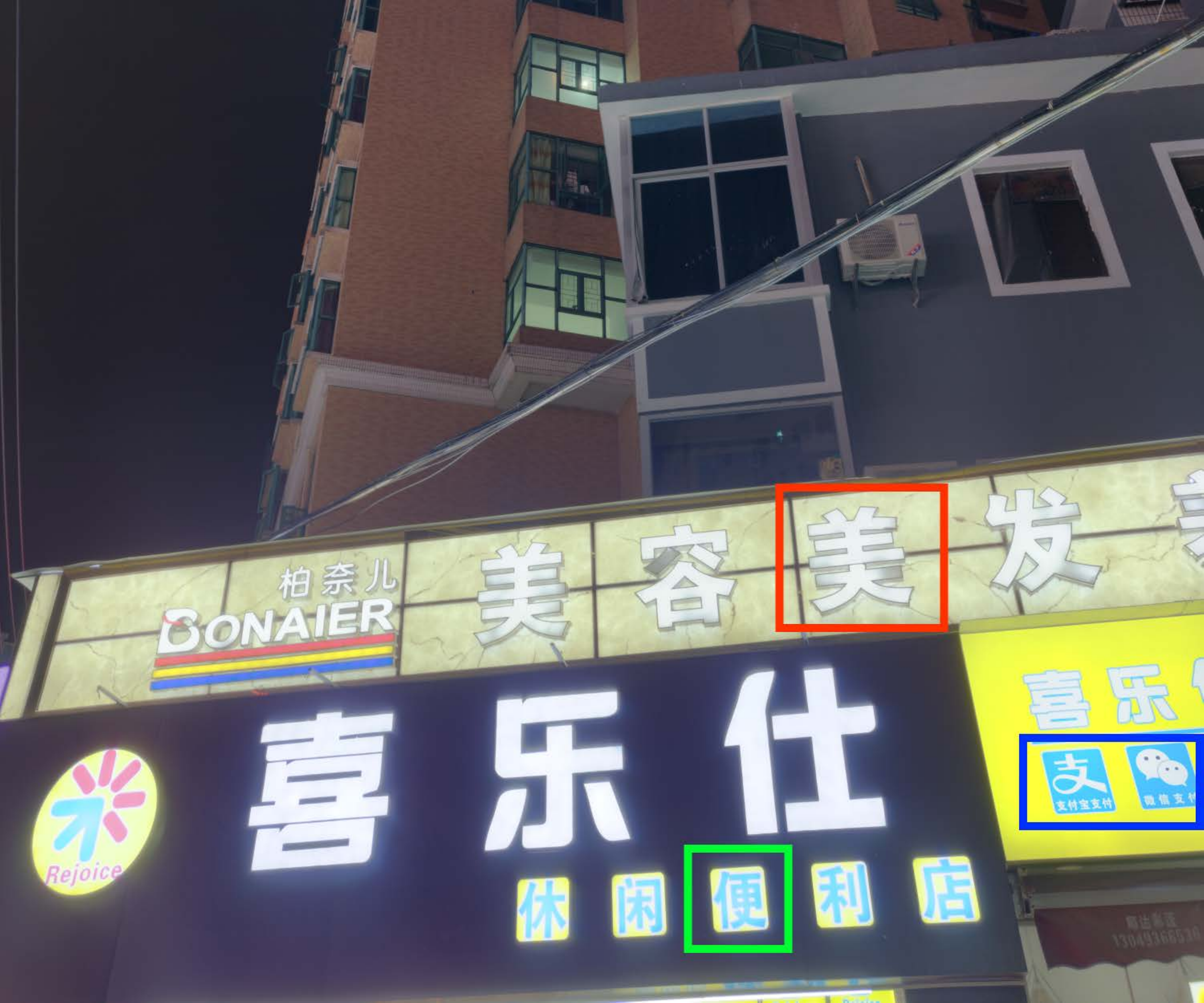} \\
                \includegraphics[width=\swthree,angle=0]{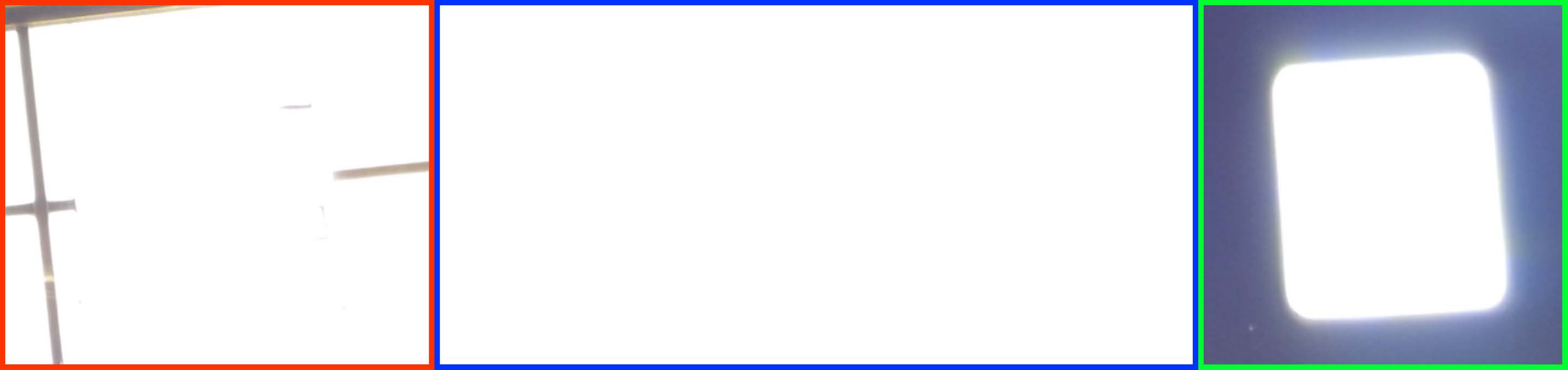} &
                \includegraphics[width=\swthree,angle=0]{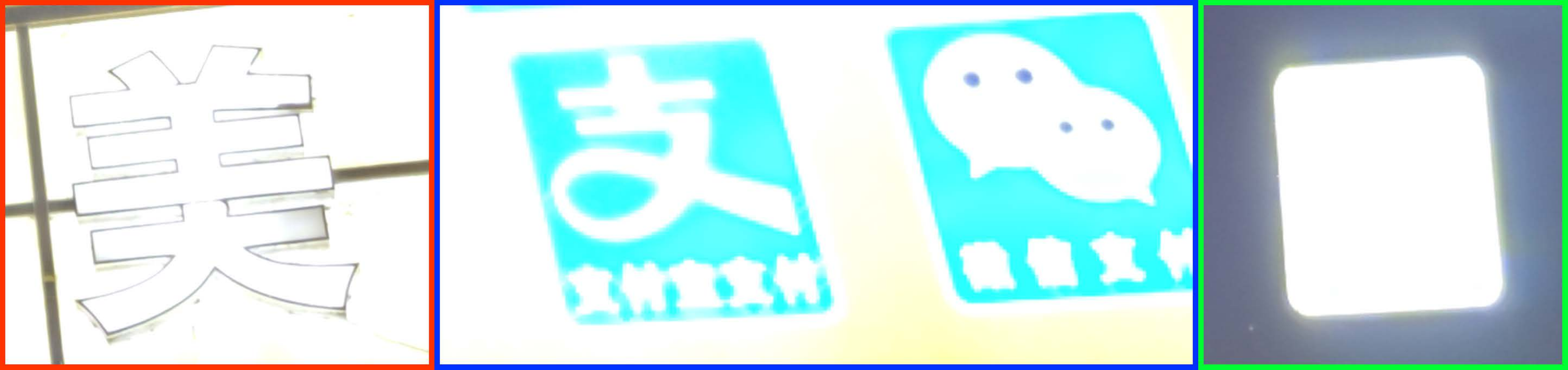} &
                \includegraphics[width=\swthree,angle=0]{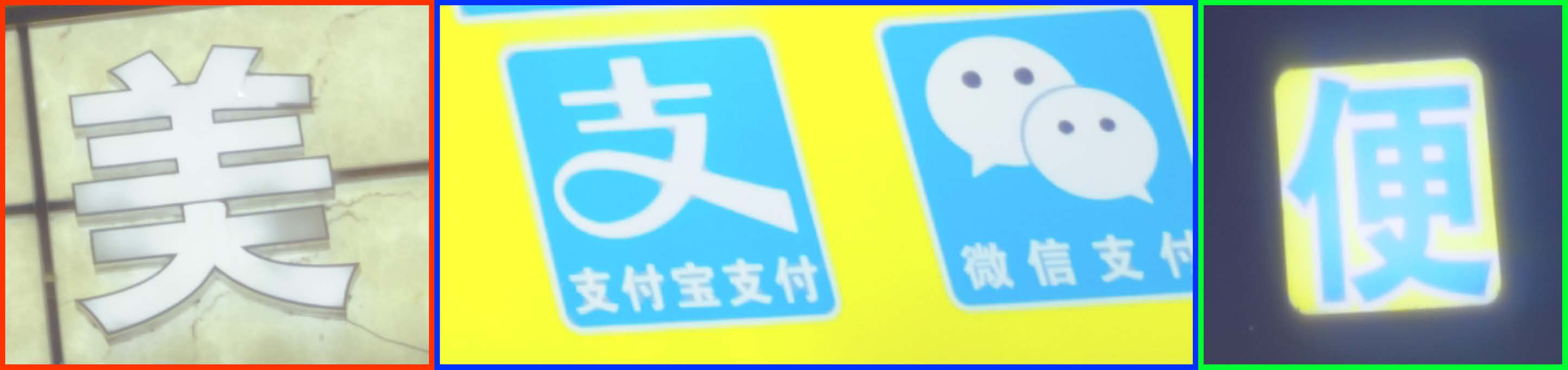} \\
                (a) AE preview $\mathbf{x}$ & (b) Pourreza-Shahri~\cite{pourreza2015exposure} & (c) Ours \\
                exposure bracketing & $\{\mathbf{z}_0, \mathbf{z}_5\}$ & $\{\mathbf{z}_1, \mathbf{z}_4, \mathbf{z}_6\}$ \\
            \end{tabular}
            \end{center}
            \vspace{-3mm}
            \caption{
            (a) is an AE preview image while (b) and (c) are the HDR images generated with the exposure bracketing selection methods proposed by Pourreza-Shahri et al.~\cite{pourreza2015exposure} and us.
            Both methods implement exposure bracketing selection only based on the AE preview image.
            Our proposed method can attain more details in saturated areas than ~\cite{pourreza2015exposure}.
            Since Pourreza-Shahri et al.~\cite{pourreza2015exposure} focus on the illumination of the AE image, it does not know what the true brightness is in the saturated areas.
            However, our method considers the illumination and semantic information of the AE scene at the same time.
                        Though the area is saturated, it is still possible to be recognized as billboard with semantic information.
            Therefore, our method tends to choose the exposure bracketing ($\{\mathbf{z}_1, \mathbf{z}_4, \mathbf{z}_6\}$) that captured the details of the billboard.
            }
            \label{fig:figure1}
            \end{figure}
            
            Most of the existing exposure bracketing selection methods propose to either find a minimal-bracketing set to cover the whole dynamic range of the scene~\cite{barakat2008minimal} or to maximize the signal-to-noise ratio (SNR)~\cite{hasinoff2010noise, gallo2012metering, seshadrinathan2012noise, van2018improved}.
            Although these methods have a decent performance, they have many limitations.
            For example, Hasinoff et al.~\cite{hasinoff2010noise} and Seshadrinathan et al.~\cite{seshadrinathan2012noise} can only handle raw images that are linearly related to the scene irradiance.
            To be more flexible with both linear and non-linear operations on the camera \eg the image signal processor (ISP), Barakat et al.~\cite{barakat2008minimal}, Gallo et al.~\cite{gallo2012metering} and Beek et al.~\cite{van2018improved} pre-estimate the camera response function based on \cite{debevec2008recovering}.
            However, all these methods need to employ either the dynamic range~\cite{barakat2008minimal, hasinoff2010noise} or the full irradiance histogram~\cite{gallo2012metering, seshadrinathan2012noise, van2018improved} of the scene by using a stream of preview images.
            Even though the preview images are easy to access sometimes, it is not user-friendly to preview the scene with different exposures.
            To alleviate this issue, Huang et al.~\cite{huang2013intelligent} and Pourreza-Shahri et al.~\cite{pourreza2015exposure, pourreza2015automatic} select exposures according to the scene information from the auto-exposure (AE) preview image and the camera parameters.
            As a result, they fail to capture information beyond the dynamic range from this single preview image, leading to sub-optimal performance.

            Although neural networks are widely used to generate HDR image~\cite{eilertsen2017hdr, yang2018image, wu2018deep, yan2019attention, prabhakar2017deepfuse, cai2018learning}, they have not been applied to select exposure bracketing.
            The proposed exposure bracketing selection network (EBSNet) can extract both illumination as well as semantic information from a single low-resolution LDR image.
            Unlike \cite{huang2013intelligent, pourreza2015exposure, pourreza2015automatic} only use illumination information from a single LDR image, the proposed EBSNet can utilize the information beyond the dynamic range of the single LDR image, such as the saturated light and the indistinguishable details in the dark grassland by considering the semantic information.
            Figure~\ref{fig:figure1} shows an example whose input image is over-exposure in the billboard and the method proposed by Pourreza-Shahri et al.~\cite{pourreza2015exposure} fails to predict an exposure bracketing to cover all the details in these areas.
            However, with the analysis of semantic information and illumination distribution of the preview image, our model succeeds in finding the exposure that can capture the content under the billboard which is always informative.
            In addition, for some scenarios such as mobile applications where the raw data, the camera response function or the sensor noise model parameters are not accessible, the existing exposure bracketing selection methods \cite{barakat2008minimal, hasinoff2010noise, gallo2012metering, seshadrinathan2012noise, huang2013intelligent, pourreza2015exposure, pourreza2015automatic, van2018improved} that strongly rely on the above additional information are infeasible.
            %
            However, the proposed EBSNet considers less of these information and can automatically learn to select out the appropriate exposure bracketing.
            %
            
            %
            We employ Reinforcement Learning~\cite{sutton1998introduction} to update the parameters in EBSNet.
            In this paper, we propose to train the proposed EBSNet as an agent, which is rewarded by the output of the multi-exposure fusion against the ground truth HDR images.
            Specifically, if the currently selected exposure bracketing generates a higher quality HDR image, the EBSNet will get a positive reward, and vice versa.
            %
            %
            We propose a multi-exposure fusion network (MEFNet) to merge the images of different exposures predicted by EBSNet and generate an HDR image. In the meantime, it will feedback a reward to EBSNet for the update of EBSNet.
            EBSNet and MEFNet can be trained jointly and the entire framework can produce favorable results against recent state-of-the-art methods.

            To facilitate future research, we also collect a dataset.
            Our dataset contains 800 samples of various scenes, such as indoor and outdoor scenes, night sight and day sight scenes.
            For each scene, it consists of an image obtained under auto-exposure and ten images whose exposure time are related to the auto-exposure image.
            Specifically, the image captured by using auto-exposure is used to determine the optimal exposure bracketing, while the remaining ten images with different exposures are the candidate images for constructing the predicted optimal exposure bracketing.
            These ten images are also used to generate the ground truth for HDR image generation.

            The \textbf{contributions} of this work are as follows.
            \begin{itemize}
                \item 
                The proposed EBSNet can use the illumination and semantic information of a low-resolution auto-exposure image to predict the optimal exposure bracketing without knowing the camera response function as well as the sensor noise model parameters and can also adapt to the images non-linear to the scene irradiance.
                
                \item The EBSNet is trained as an agent in reinforcement learning and rewarded with a novel MEFNet. The EBSNet and MEFNet can be trained jointly and they can perform favorably against state-of-the-art exposure bracketing selection methods.
                \item We provide a new benchmark dataset to facilitate the future studies of exposure bracketing selection.
            \end{itemize}
    
        \begin{figure*}
        \begin{center}
        \includegraphics[width=0.95\linewidth]{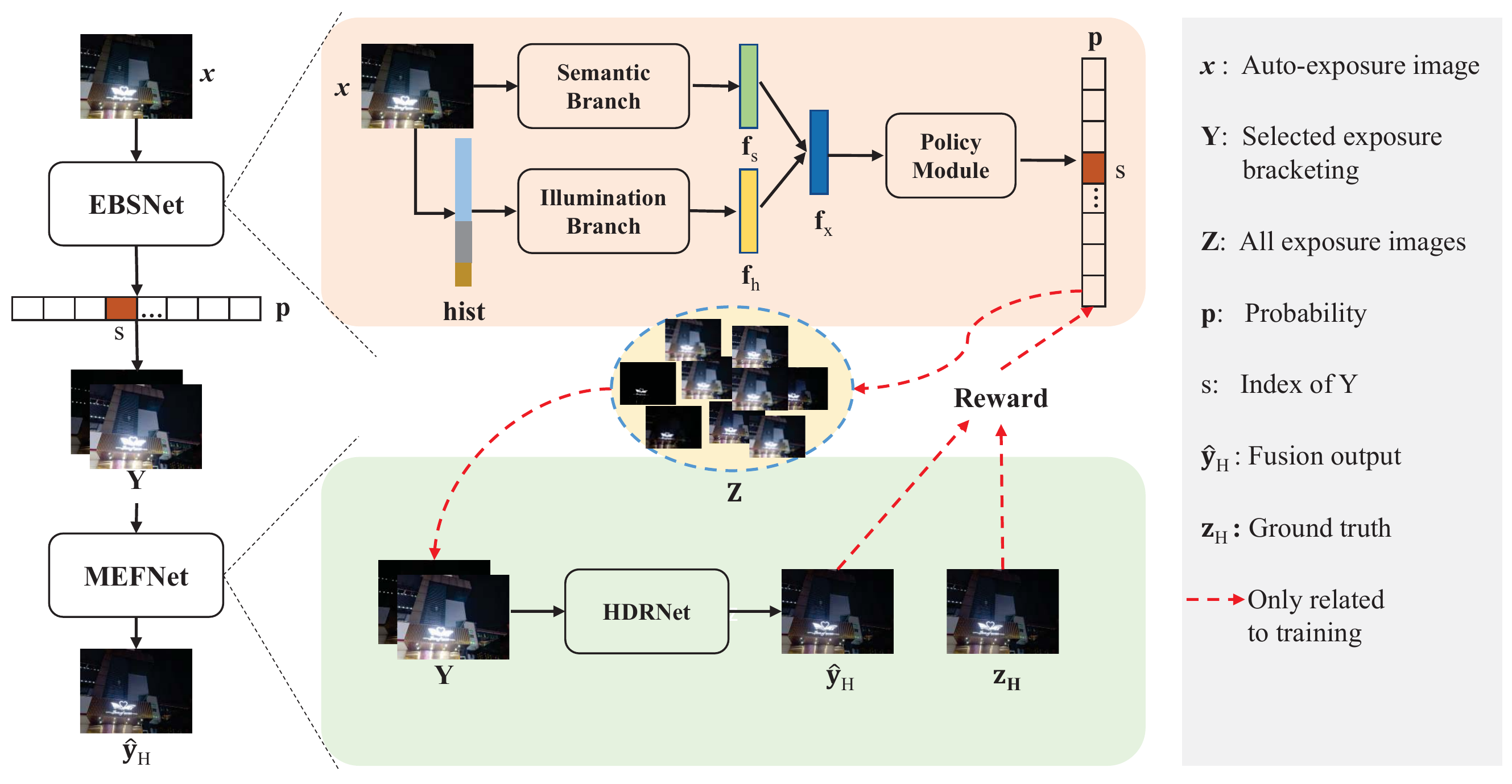}
        \end{center}
        \vspace{-3mm}
        \caption{This figure shows the pipeline of our proposed method.
        First, it captures an auto-exposure low-resolution preview image, and then feeds it into an Exposure Bracketing Selection Network (EBSNet) to determine an optimal exposure bracketing according to the extracted illumination and semantic information. 
        Multi-Exposure Fusion Network (MEFNet) will take the selected exposure bracketing as input and generate an HDR image.
        When training EBSNet, We regard it as an agent and it gets the reward from MEFNet.
        EBSNet and MEFNet can be trained jointly.
        Please see the manuscript for more details.
        } 
        \label{fig:framework}
        \end{figure*}
    
        \section{Related Work}
        \textbf{Exposure Bracketing Selection.}
        Most of the existing Exposure Bracket Selection methods are based on the estimation of the dynamic range.
        Barakat et al.~\cite{barakat2008minimal} tend to find the minimal-bracketing sets for attaining HDR images by assuming a linear camera response function and a uniform irradiance distribution.
        Hasinoff et al.~\cite{hasinoff2010noise} focus on achieving higher worst-case SNR by minimizing the number of exposures or the total exposure duration.
        Grossberg et al.~\cite{grossberg2003high} optimize the exposure bracket by minimizing an objective function that is based on the derivative of the response function.
        While the above three methods are based on the estimation of the extent of the dynamic range, Gallo et al.~\cite{gallo2012metering}, Seshadrinathan et al.~\cite{seshadrinathan2012noise} and Beek et al.~\cite{van2018improved} present a method to estimate the full irradiance histogram of the scene by capturing a stream of images and a strategy to select the set of exposures that need to be acquired. 
        Also, some methods do not require the scene irradiance to be known~\cite{huang2013intelligent, pourreza2015automatic, pourreza2015exposure} and they use the scene information from the normal or auto-exposure and the camera parameters to perform exposure bracketing.

        For releasing our model from the estimation of the dynamic range and response function, our proposed model only has a view on the auto-exposure low-resolution preview LDR image.
        Different from the similar works~\cite{huang2013intelligent, pourreza2015automatic, pourreza2015exposure} who just focus on the illumination of AE image, our method also take consideration on the semantic information of the preview image by a neural network which can utilize the information beyond the dynamic range of the AE image, such as the saturated light and the indistinguishable details in the dark grassland.

        \textbf{Multi-Exposure Fusion.}
        Traditional Multi-Exposure Fusion methods mainly use hand-crafted features.
        Mertens et al.~\cite{mertens2009exposure} fuse the images by using simple quality measures such as saturation and contrast.
        Ma et al.~\cite{ma2017robust} decompose the image patch into three components and fuse them separately to obtain the desired one.
        There are also some methods of attaining fusion weights by using image filters.
        For example, Li et al.~\cite{li2013image} decompose an image into two layers and employ a guided filtering-based weighted average technique to make full use of the spatial consistency for exposure fusion.
        Raman et al.~\cite{raman2009bilateral} use bilateral filters to preserve details in both bright and dark regions of the scene.
        As the previous methods neglect the semantic information of the images, \cite{cai2018learning,prabhakar2017deepfuse} introduce CNNs into Multi-Exposure Fusion to extract the high-level features with generated ground truth~\cite{cai2018learning} or without reference~\cite{prabhakar2017deepfuse}.
        Our MEFNet aims to provide a reward for the update of the EBSNet.
        It is easy to replace our MEFNet with the above MEF method.
        However, for modeling the interaction between the exposure bracketing selection and exposure fusion and making them can be trained jointly, we tend to implement exposure fusion with a neural network.
        
        \textbf{Reinforcement Learning.}
        Recently, reinforcement learning (RL) has been applied to all kind of computer vision tasks, such as classification~\cite{ba2014multiple, mnih2014recurrent}, detection~\cite{jie2016tree}, visual captioning~\cite{liu2017improved,li2019end,rennie2017self}, machine translation~\cite{wu2016google}, and so on. 
        There are also some exposure-related works updated via reinforcement learning \cite{yang2018personalized,yu2018deepexposure}.
        However, Yang et al.~\cite{yang2018personalized} aim to build a system that users can adapt the exposure of a single image personally via giving a score to a preview image and Yu et al.~\cite{yu2018deepexposure} use RL to learn local exposures for exposure fusion with segmentation and carefully retouch.
        Our model tends to use RL to determine the optimal exposure bracketing for the generation of an HDR image by analyzing the preview image and get a reward directly from a multi-exposure fusion network.
        By joint training, our EBSNet and MEFNet can get benefit from each other.

        \section {Method}

        The purpose of this method is to select an optimal exposure bracketing for HDR image generation and it contains two components: EBSNet and MEFNet which are showed in Figure~\ref{fig:framework}.
        Unlike most of the existing methods that need to consider a sequence of preview images with different exposures, the proposed EBSNet only needs to analyze a single AE low-resolution preview image by considering both the illumination distribution and the semantic information.
        %
        Since we cannot directly get the ground truth of the exposure bracketing, we propose to update EBSNet via RL~\cite{sutton1998introduction} guided by a reward generated from exposure fusion and its HDR ground truth.
        Even though any exposure fusion method can be used in RL framework to get the reward, we propose a network MEFNet to fuse the selected exposure bracketing since both EBSNet and MEFNet can benefit from each other through joint training.
        
        \subsection{Exposure Bracketing Selection Network (EBSNet)}
        The input of the proposed EBSNet is a low-resolution AE LDR preview image $\mathbf{x}$ and EBSNet predicts a bracketing of $K$ different exposures $\mathbf{Y}=\{\mathbf{y}_0, \mathbf{y}_1, \cdots, \mathbf{y}_{K-1}\}$ which may cover the dynamic range of the scene as large as possible.
        Since there is only one LDR image $\mathbf{x}$, it is difficult to select out an optimal exposure bracketing only based on the illumination distribution of $\mathbf{x}$ especially in the area where is over-exposure or extremely dark.
        For a better selection, EBSNet takes the semantic information of the preview image into consideration and it is favorable.
        As Figure~\ref{fig:figure1} shown, though the billboard in the AE image is saturated and its illumination information has been discarded, it is still possible to be recognized as a billboard with its semantic information.
        Since billboards are used for displaying informative content, our model tends to get an exposure bracketing that captured the details of the billboard.
        Therefore, the proposed EBSNet consists of two branches.
        One is \textbf{semantic branch} for semantic feature extraction while the other is \textbf{illumination branch} for illumination feature extraction (as shown in Figure~\ref{fig:framework}).

        \paragraph{Semantic Branch}
        The semantic feature extraction is constructed as Alexnet \cite{krizhevsky2012imagenet}.
        It takes the preview image $\mathbf{x}$ as input and extracts the semantic feature $\mathbf{f_s}$ from the first fully-connected layer. We can formulate it as:
        \begin{equation}
            \mathbf{f_s} = \mathrm{Alexnet}(\mathbf{x}).
        \end{equation}

        \paragraph{Illumination Branch}
        In order to utilize the illumination distribution globally and locally, we estimate the histogram of $\mathbf{x}$ with a 3-levels spatial pyramid.
        For $u$-th level, we divide $\mathbf{x}$ into $2^{(u-1)}\times2^{(u-1)}$ non-overlapped patches and a histogram is estimated from every patch with 32 bins.
        By concatenating all the histograms, we get feature $\mathbf{hist}$ and send it to a neural network that consists of two convolutional layers and one fully-connected layer.
        Finally, we attain the illumination feature $\mathbf{f_h}$. 
        By denoting this neural network as Histnet, we can represent $\mathbf{f_h}$ as:
        \begin{equation}
            \mathbf{f_h} = \mathrm{Histnet}(\mathbf{hist}).
        \end{equation}

        To fuse these two features, we introduce an additional fully-connected layer which takes the concatenation of $\mathbf{f_s}$ and $\mathbf{f_h}$ as inputs and gets the final scene feature $\mathbf{f_x}$ of preview image $\mathbf{x}$ which is shown as:
        \begin{equation}
            \mathbf{f_x} = \mathrm{fc}([\mathbf{f_s},  \mathbf{f_h}]).
        \end{equation}
        %
        
        %
        \paragraph{Policy Module}
        Similar to the existing methods, e.g. \cite{barakat2008minimal, hasinoff2010noise}, the exposure $\mathbf{y}_k$ ($k \in [0, K-1]$) in the selected optimal exposure bracketing $\mathbf{Y}$ is included in a large exposure bracketing $\mathbf{Z}=\{\mathbf{z}_0, \mathbf{z}_1, \cdots, \mathbf{z}_{J-1}\}$ which has $J$ different exposures and contains a large dynamic range.
        %
        In this way, there are totally $N$=$C_J^K$ possible exposure bracketing in a scene and the candidate exposure bracketing is denoted as $\mathbf{A}=\{\mathbf{Y}_0, \mathbf{Y}_1, \cdots, \mathbf{Y}_{N-1}\}$.

        Given the scene feature $\mathbf{f_x}$, a policy module is used to predict a probability distribution ${\mathbf{p}}$ over the candidate exposure bracketing $\mathbf{A}$ and the exposure bracketing $\mathbf{Y}_{s}$ with the highest probability will be sent into MEFNet to further generate the HDR image.
        The policy module can be described as follows:
        \begin{equation}
            \begin{aligned}
                \mathbf{p}&=\mathrm{Softmax}(\mathbf{Wf_x} + \mathbf{b}) \\
                s&=\mathrm{argmax}(\mathbf{p}),
                \label{equ:decision}
            \end{aligned}
        \end{equation}
        where $\mathbf{W}$ and $\mathbf{b}$ are the weights and bias of a fully-connected layer, and $s$ is the index of highest probability in $\mathbf{p}$.
        
        \begin{figure*}[t]\footnotesize
        \begin{center}
        \begin{tabular}{cccccc}
        \includegraphics[width=\swseven,angle=0]{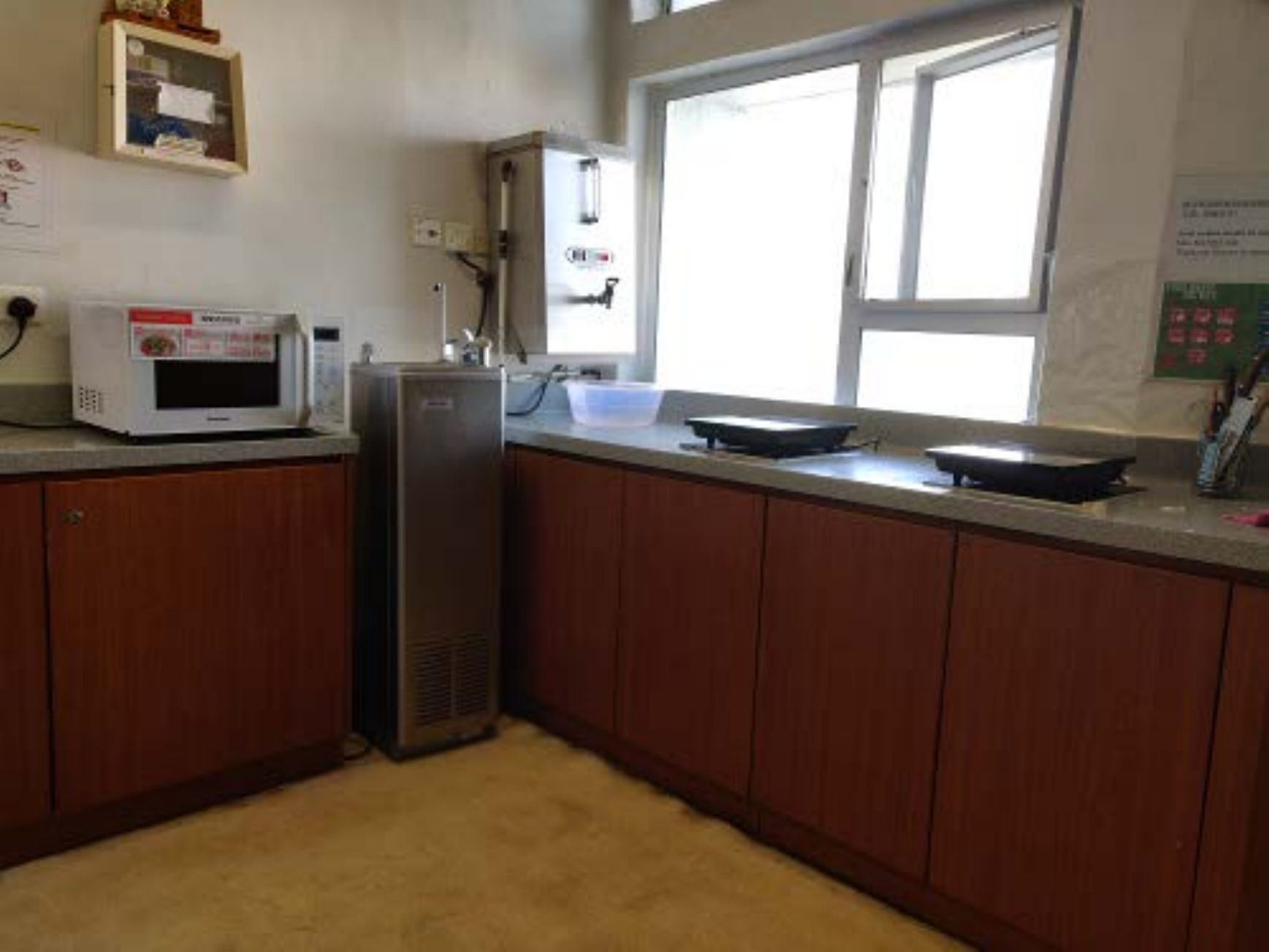} &
        \includegraphics[width=\swseven,angle=0]{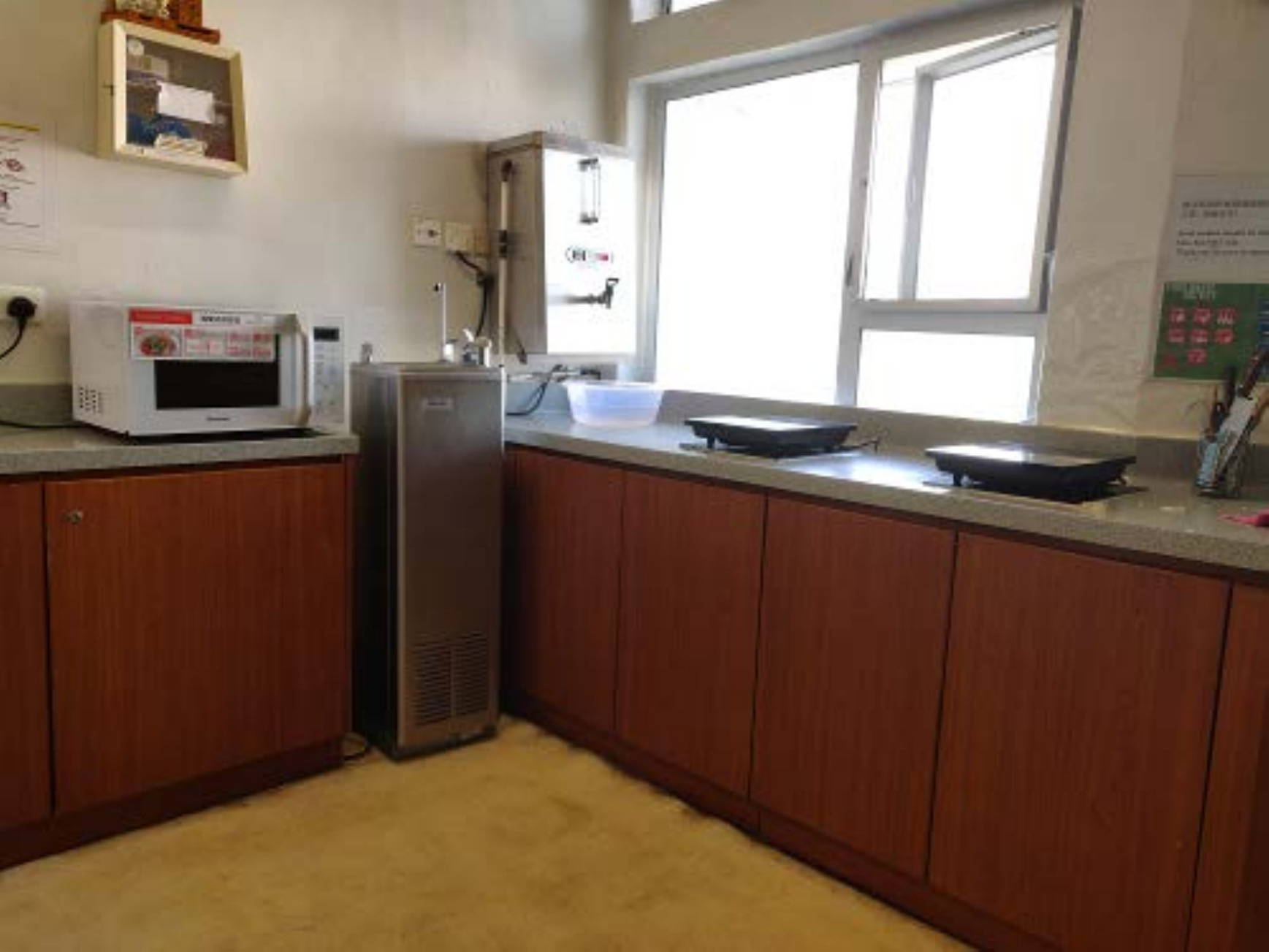} &
        \includegraphics[width=\swseven,angle=0]{figures/samples/step_2.pdf} &
        \includegraphics[width=\swseven,angle=0]{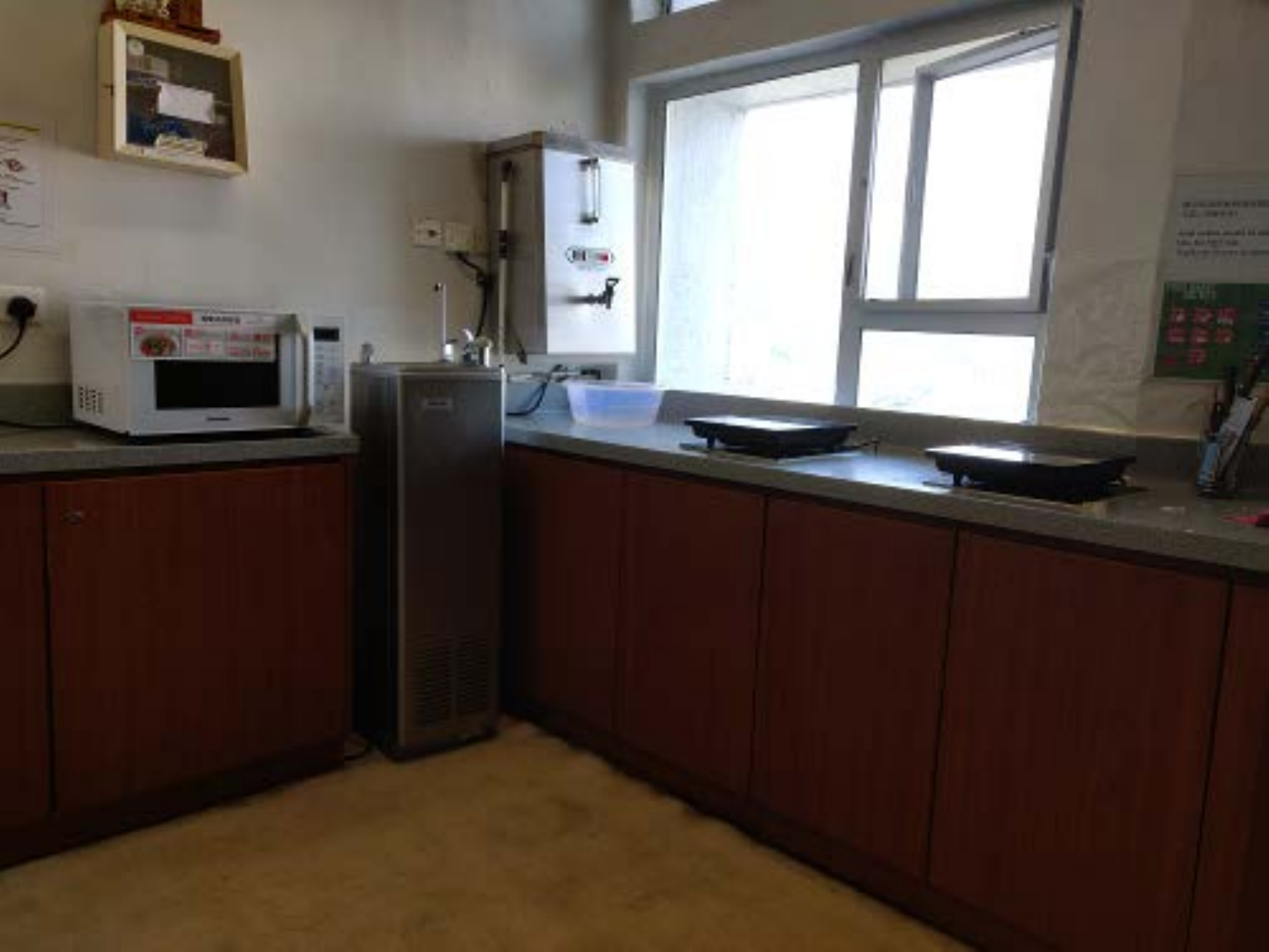}&
        \includegraphics[width=\swseven,angle=0]{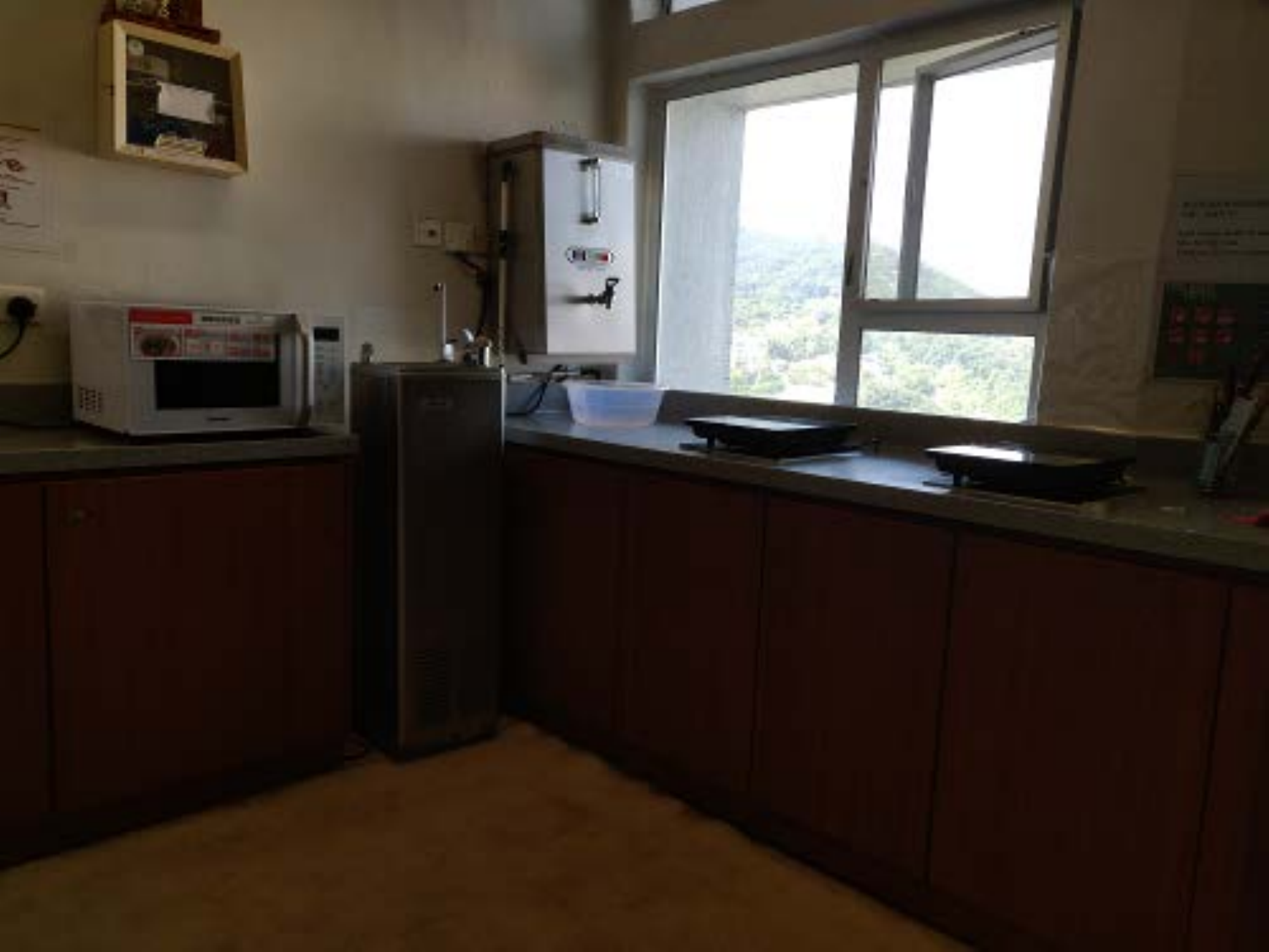} &
        \includegraphics[width=\swseven,angle=0]{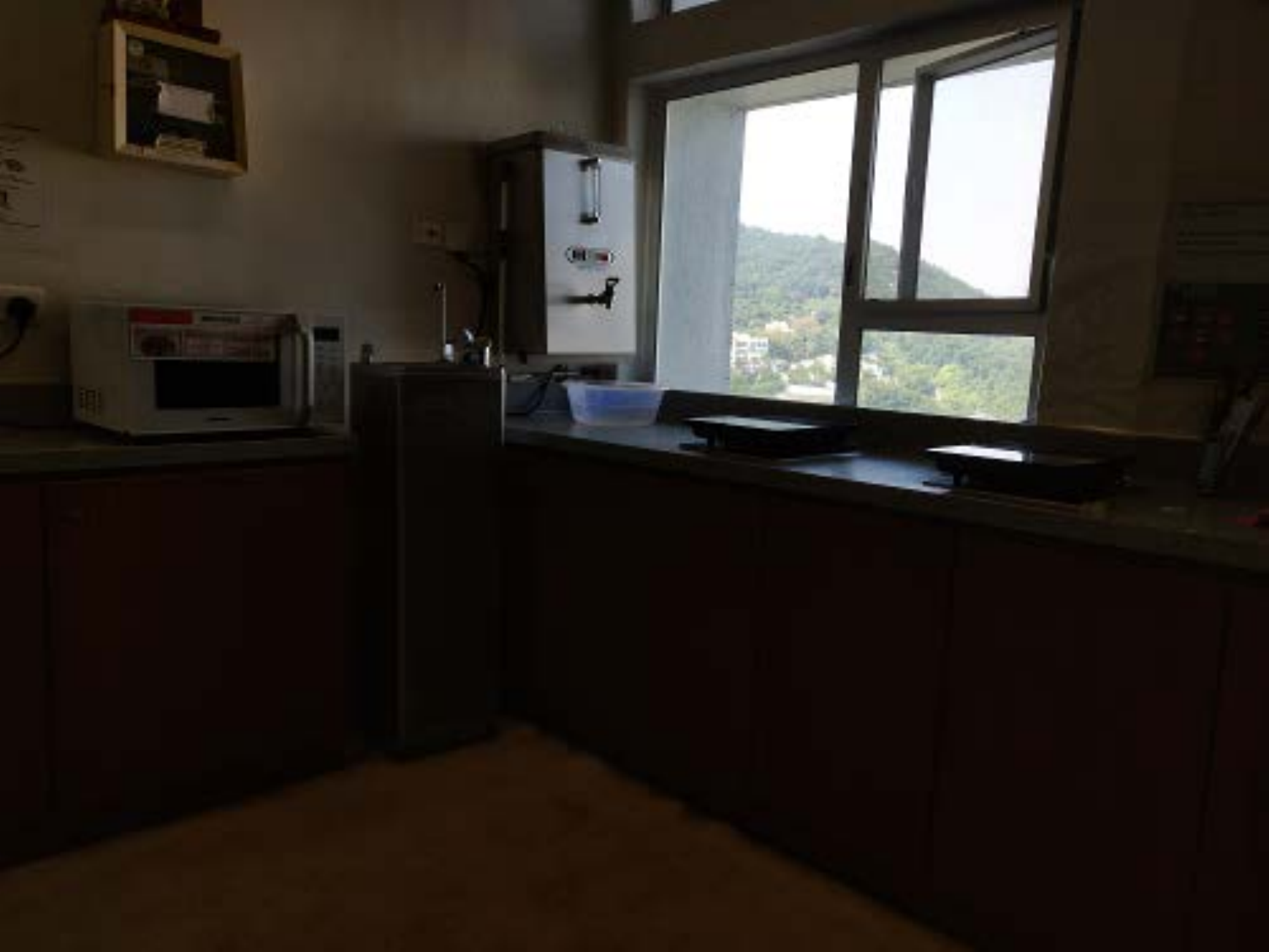} \\
        $\mathbf{x}$:13.4ms&$\mathbf{z}_0$:26.8ms&$\mathbf{z}_1$:20.1ms&$\mathbf{z}_2$:13.4ms&$\mathbf{z}_3$:6.7ms&$\mathbf{z}_4$:3.35ms \\
        
        \includegraphics[width=\swseven,angle=0]{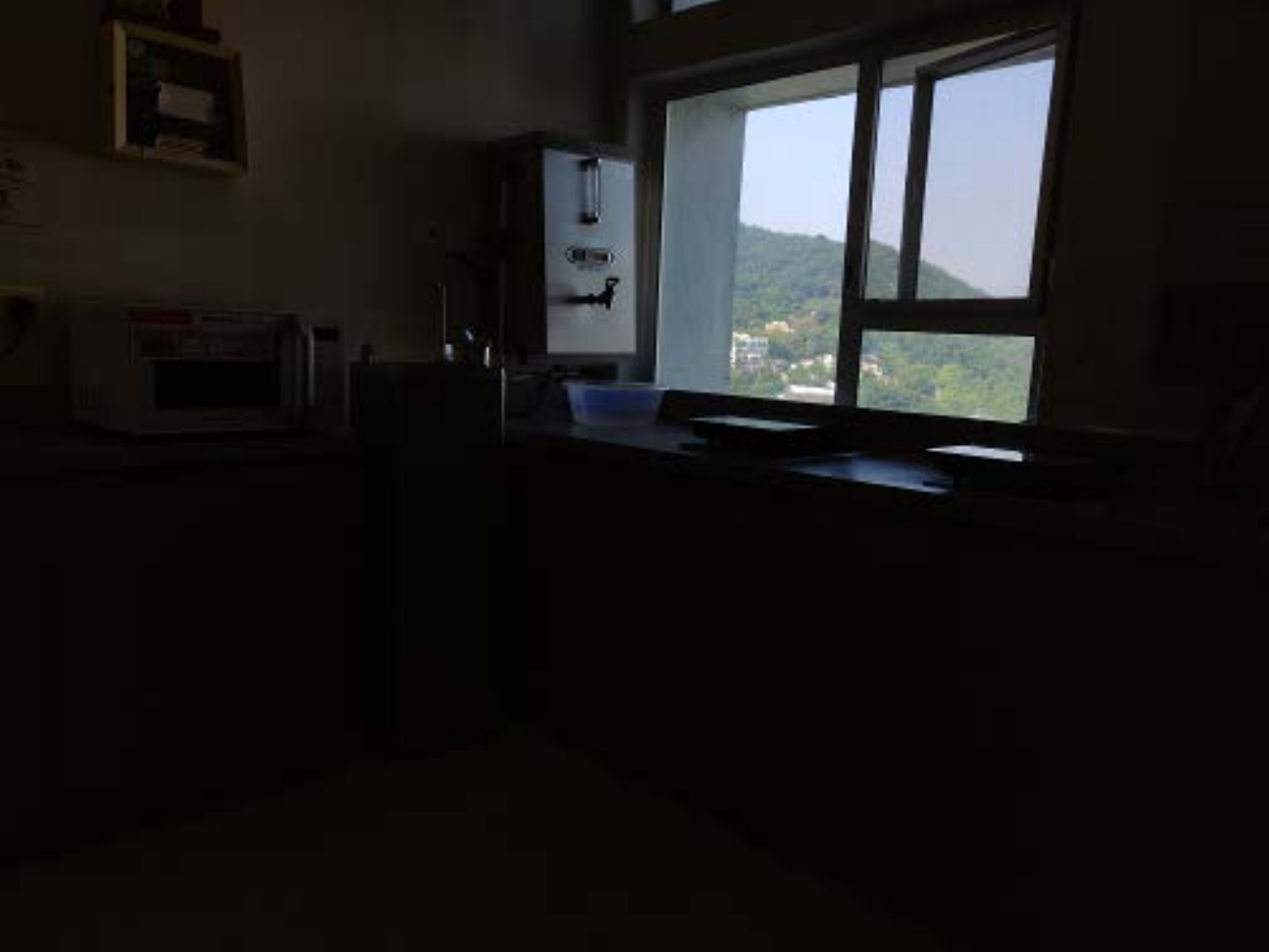} &
        \includegraphics[width=\swseven,angle=0]{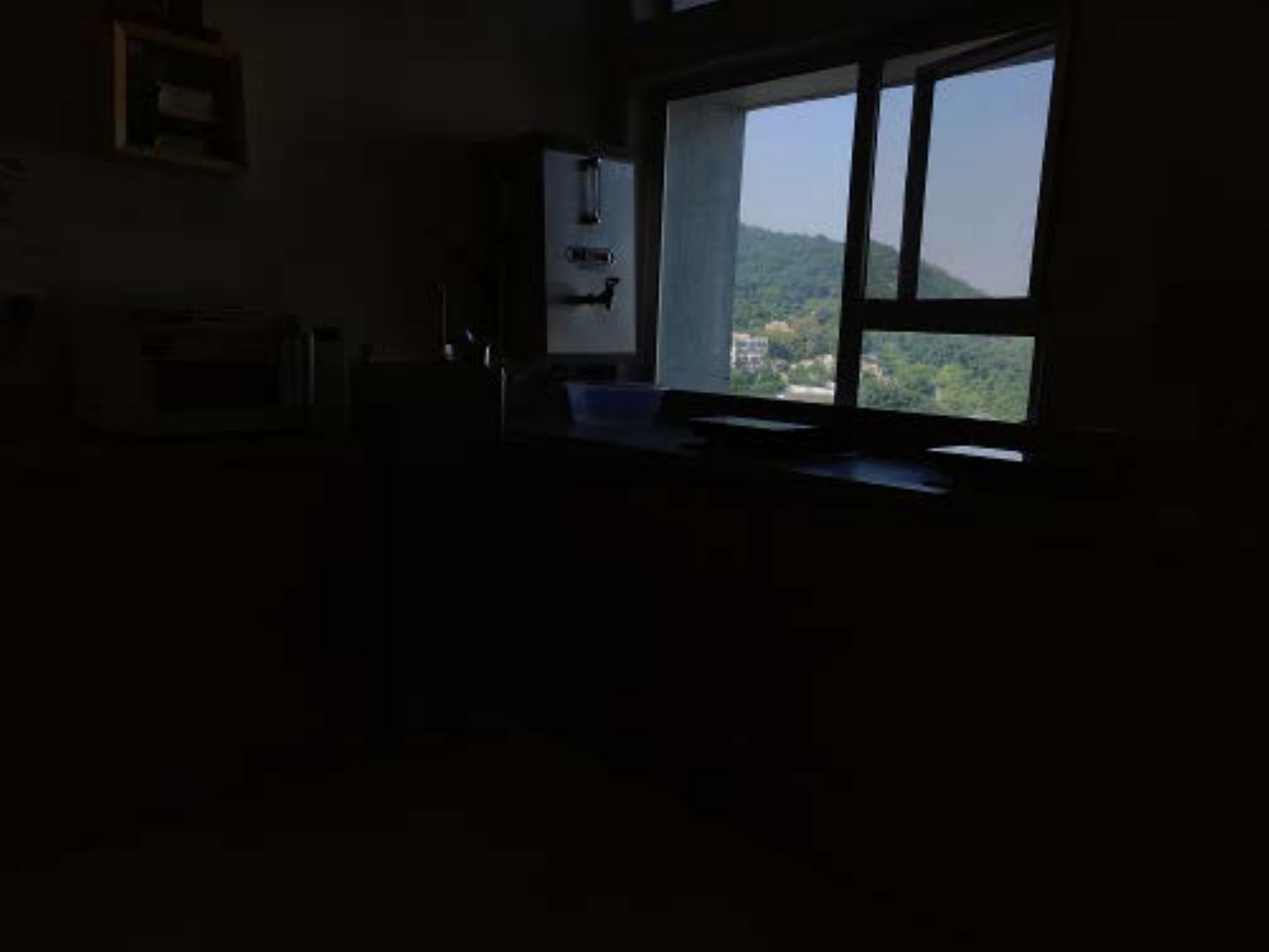} &
        \includegraphics[width=\swseven,angle=0]{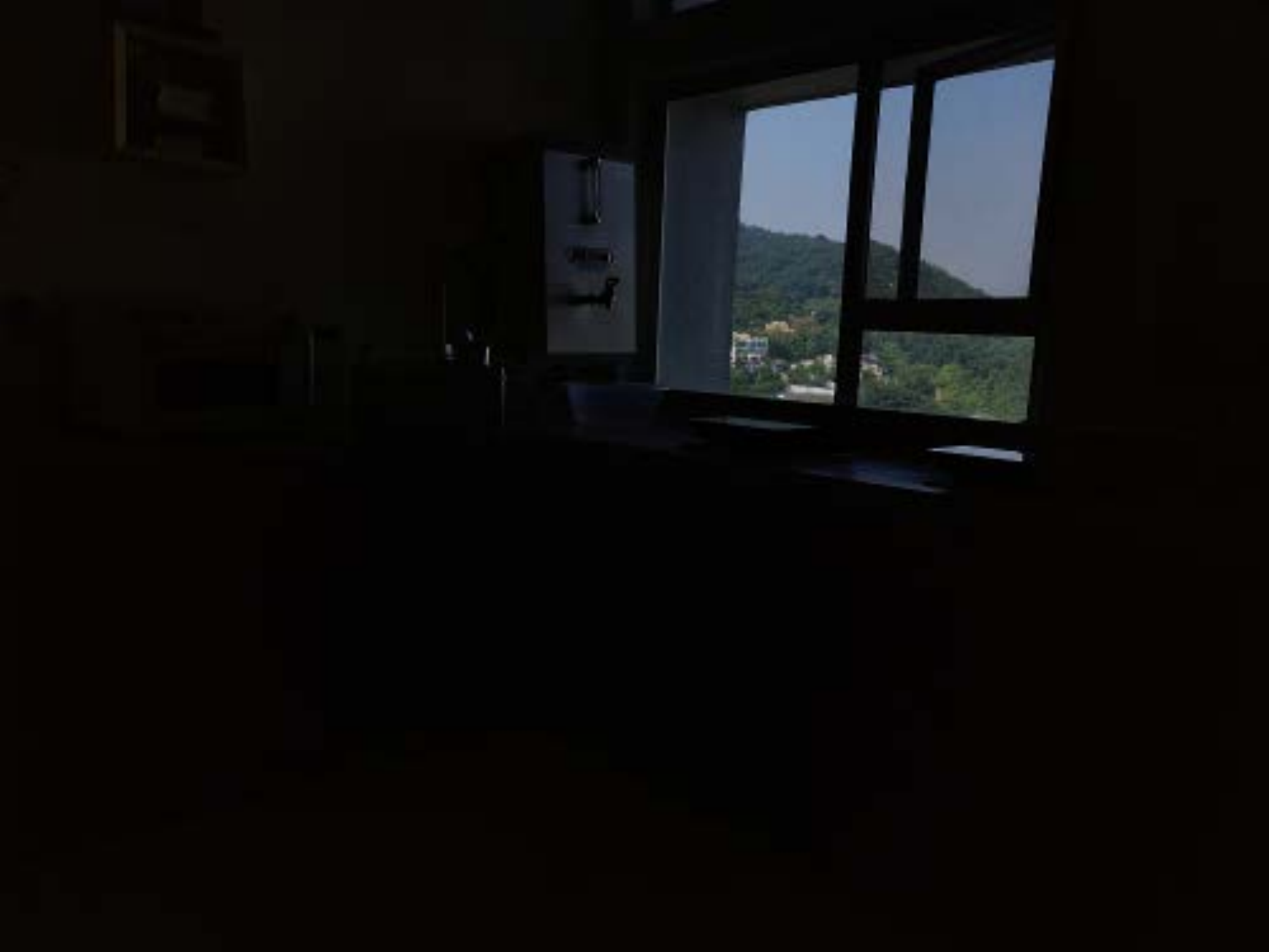} &
        \includegraphics[width=\swseven,angle=0]{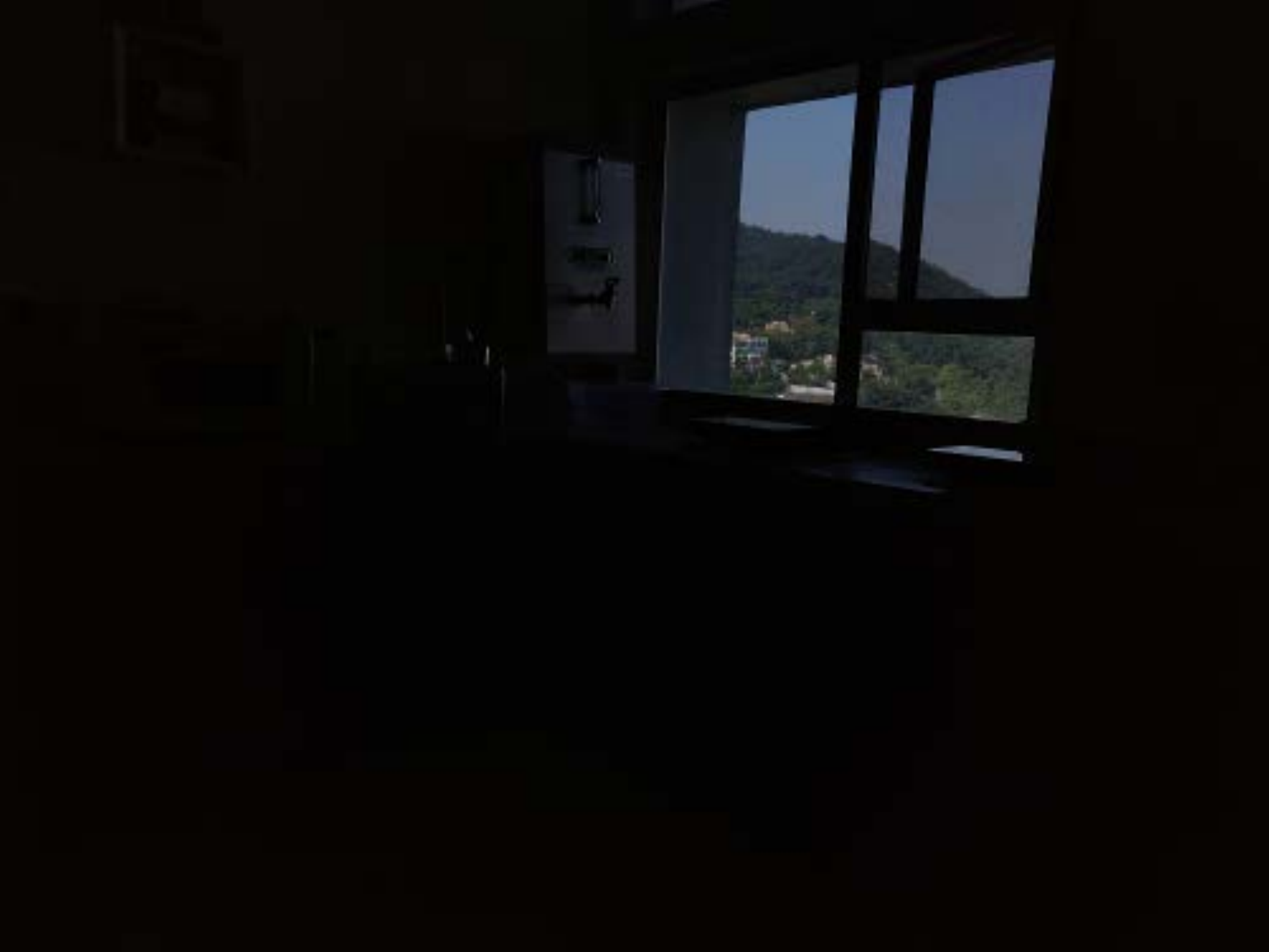} &
        \includegraphics[width=\swseven,angle=0]{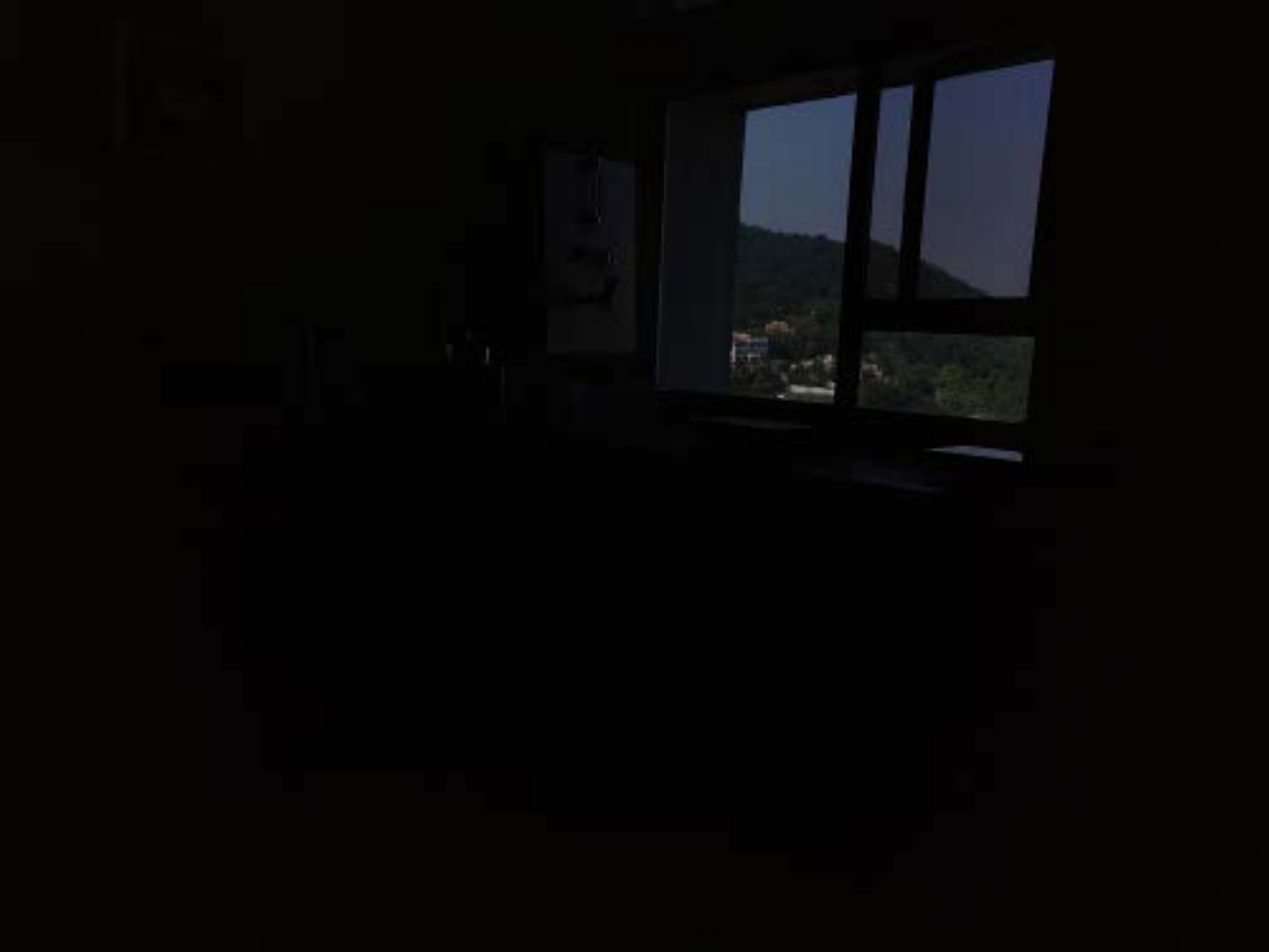} &
        \includegraphics[width=\swseven,angle=0]{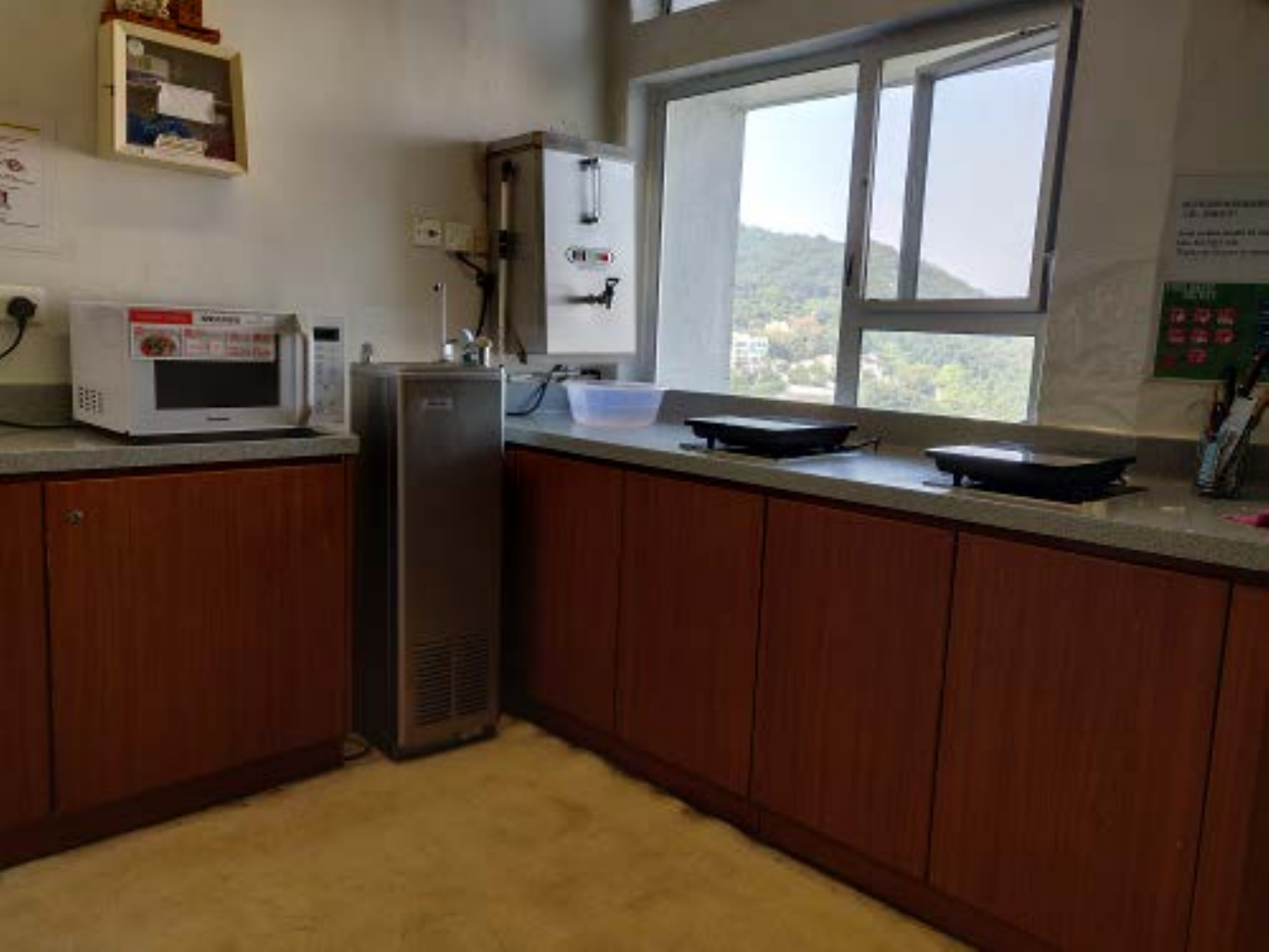} \\
        $\mathbf{z}_5$:1.68ms&$\mathbf{z}_6$:0.84ms&$\mathbf{z}_7$:0.42ms&$\mathbf{z}_8$:0.21ms&$\mathbf{z}_9$:0.10ms&$\mathbf{z_H}$
        
        \end{tabular}
        \end{center}
        \vspace{-3mm}
        \caption{
        One sample in our dataset. 
        It consists of an auto-exposure image $\mathbf{x}$ and ten images $\{\mathbf{z}_0, \mathbf{z}_1, \cdots, \mathbf{z}_9\}$ whose exposure time are related to the auto-exposure. 
        %
        The exposure durations are also shown here.
        Best viewed in color; zoom in for additional details.
        }
        \label{fig:samples}
        \end{figure*}
        
        \subsection{Multi-Exposure Fusion Network (MEFNet)}
        By considering the $K$ LDR images in $\mathbf{Y}_{s}$ selected by EBSNet, exposure fusion can generate an HDR image $\mathbf{\hat{y}}_H$.
        Also, exposure fusion is used to update EBSNet.
        %
        Since there is no ground truth for the selection procedure, we propose to use RL scheme for the training of EBSNet. 
        With RL, the agent (EBSNet) will decide the action (one of the exposure bracketings in $\mathbf{A}$) in an environment (multi-exposure fusion) to maximize the cumulative reward according to the current state (quality of generated HDR image).
        
        Although a differentiable environment is not necessary for RL framework, we still adopt a learnable neural network MEFNet derived from the HDRNet~\cite{gharbi2017deep} for multi-exposure fusion.
        In this way, both EBSNet and MEFNet can benefit from each other through joint training.
        MEFNet can be expressed as:
        \begin{equation}
            \mathbf{\hat{y}}_H = \mathrm{HDRNet}(\mathbf{Y}_{s}).
        \end{equation}

        \subsection{Learning}
        \paragraph {MEFNet loss}
        When training MEFNet, it minimizes the Charbonnier Loss~\cite{barron2019general} which is defined as:
        \begin{equation}
            \mathcal{L}_f(\mathbf{\hat{y}}_H, \mathbf{z}_H)=\sum_i\sum_j\sqrt{(\mathbf{\hat{y}}_H(i,j) - \mathbf{z}_H(i,j))^2+\epsilon^2},
            \label{equ: c-loss}
        \end{equation}
        where $(i,j)$ is the position in the image and $\epsilon=10^{-3}$.
        To generate the ground truth HDR image $\mathbf{z}_H$, we fuse all images in $\mathbf{Z}$ with an existing exposure fusion method~\cite{mertens2009exposure}.

        \paragraph{EBSNet Loss}
        In RL, the agent needs to gather as much reward as possible.
        In this paper, the reward is defined as the difference between the current and previous peak signal-to-noise ratio (PSNR) of the HDR image as:
        \begin{equation}
            \mathcal{R}(\mathbf{\hat{y}}_H^{v-1}, \mathbf{\hat{y}}_H^v, \mathbf{z}_H)=\mathrm{PSNR}(\mathbf{\hat{y}}_H^v, \mathbf{z}_H)-\mathrm{PSNR}(\mathbf{\hat{y}}_H^{v-1}, \mathbf{z}_H),
            \label{equ: reward}
        \end{equation}
        in which $\mathbf{\hat{y}}_H^v$ is the estimated HDR image generated from the selected exposure bracketing in $v$-th step.
        To update EBSNet, we need to minimize the following loss function as:
        \begin{equation}
            \begin{aligned}
                \mathcal{L} &= -\mathbb{E}_s[\mathcal{R}(\mathbf{\hat{y}}_H^{v-1}, \mathbf{\hat{y}}_H^v, \mathbf{z}_H)]
                \label{equ: scorefunction}
            \end{aligned}
        \end{equation}
        and
        \begin{equation}
            \begin{aligned}
                \mathbf{\nabla}\mathcal{L} &= -\mathbb{E}_s[\nabla\log \mathbf{p}(s) \mathcal{R}((\mathbf{\hat{y}}_H^{v-1}, \mathbf{\hat{y}}_H^v, \mathbf{z}_H))],
                \label{equ: gradscorefunction}
            \end{aligned}
        \end{equation}
        where $s$ is the action chosen by EBSNet represented a type of exposure bracketing and $\mathbb{E}_s[\cdot]$ is the expectation over $s$.
        %
    
         \subsection{Training Procedure}
        Since the updating of EBSNet is determined by the reward feedbacked from the MEFNet and the learning of MEFNet is affected by the selection of EBSNet, an appropriate weight initialization of both networks is necessary. 
        Therefore, before jointly training these two networks, we train them separately to get pretrain models.
        The training procedure includes three steps.
        First, we train MEFNet with $K$ images that randomly selected from the candidate exposure bracketing \textbf{A} for a more reliable reward to update EBSNet.
        Second, we use the trained MEFNet to generate HDR images and calculate the PSNRs w.r.t. the ground truth HDR image.
        The candidate exposure bracketing that has the top 1 PSNR among all the candidate exposure bracketings is set as the initial target action of the scene and is used for the coarse training of EBSNet.
        Third, we train EBSNet and MEFNet iteratively by every 10 epochs with the pretrained model attained in step one and step two.
        At that time, the input of MEFNet is selected by EBSNet, while the EBSNet is updated by the PSNR of the output of MEFNet and the ground truth HDR image.

    \section{Experiments and Analysis}
        \subsection{Dataset}
        When training the proposed framework, an auto-exposure LDR preview image as well as a stream of images with different exposures, which cover a large dynamic range, are needed. 
        As there lacks a dataset that satisfies these requirements, we propose a new dataset that contains about 800 different scenes including day and night as well as indoor and outdoor.
        Specifically, each scene includes one auto-exposure preview image $\mathbf{x}$, $J=10$ images $\mathbf{Z}$ captured under different exposures, and one HDR image $\mathbf{Z}_H$ generated by fusing all the ten images with an existing exposure fusion method~\cite{mertens2009exposure}.
        The data are collected by a mobile hold on a tripod to avoid the motion between different exposures.
        The exposure time of the AE preview denotes as $t_a$ and that of the rest ten images denote as $ 2t_a, 1.5t_a, t_a, 2^{-1}t_a, 2^{-2}t_a, 2^{-3}t_a, 2^{-4}t_a, 2^{-5}t_a, 2^{-6}t_a, \\ 2^{-7}t_a$ respectively.
        The rest of the camera settings are fixed for each exposure in a scene.
        The corresponding images are named as $\left\{ \mathbf{z}_0, \mathbf{z}_1, ..., \mathbf{z}_9 \right\}$ and one of the samples is shown in Figure~\ref{fig:samples}.

        The dataset is randomly split into 3 parts: a training set contains 600 images, a validation set contains 100 images, and a test set contains 100 images.

        \subsection{Experiment Settings}
        \paragraph{Implementation details}
        In this work, the length of each selected bracketing $K$ is set to 3 as Barakat et al.~\cite{barakat2008minimal} have shown that three images can capture the luminance of a scene adequately in most of the cases.
        Then, the total number of possible selected exposure bracketing is $N=C_{10}^3=120$.
        As some exposure bracketing are almost impossible to yield an acceptable HDR image, e.g. $\{\mathbf{z}_7, \mathbf{z}_8, \mathbf{z}_9\}$ which are too dark to capture any details in the dark regions, we only consider $N'=36$ exposure bracketing in the candidate exposure bracketing $\mathbf{A}$ in EBSNet.
        Please see the supplemental material for more details.

        While training, the input size of EBSNet and MEFNnet are $224 \times 224$ and $512 \times 512$ respectively.
        For data augmentation, we randomly crop, flip and rotate the images.
        The proposed network is trained with Adam~\cite{kingma2014adam}.
        The learning rates of MEFNet in the first step and EBSNet in the second step are set to $10^{-3}$ and $10^{-4}$ respectively.
        While in the third step, their learning rate is multiplied with $0.1$.
        The batch size is 8.

        \begin{table*}[!t]
        \begin{center}
           \begin{tabular}{|l|c|c|c|c|c|c|}
               \hline
               Method & Barakat~\cite{barakat2008minimal}  & Beek~\cite{van2018improved} & Pourreza-Shahri~\cite{pourreza2015exposure} & EBSNet+EF & EBSNet+MEFNet \\
               \hline\hline\hline
               number & 3.12 & 3.71 & 2.48 & 3.00 & 3.00 \\
               
               \hline
           \end{tabular}
        \end{center}
        \caption{ 
           Average number of selected exposure bracketing of each method.
           The number of images selected from \cite{barakat2008minimal} and \cite{van2018improved} varys from scene to scene and the maximum number of images selected from \cite{pourreza2015exposure} is 3.
           In the proposed framework, the number of the selected images $K$ is fixed to 3.
           Please see Sec.~\ref{Sec:K} for more analysis of $K$ of the proposed method in details.
        }
        \label{tb:number}
        \end{table*}

        \begin{figure*}[htp]\footnotesize
        \begin{center}
        \renewcommand{\tabcolsep}{1pt}
        \begin{tabular}{ccc}
        \includegraphics[width=\swthree,angle=0]{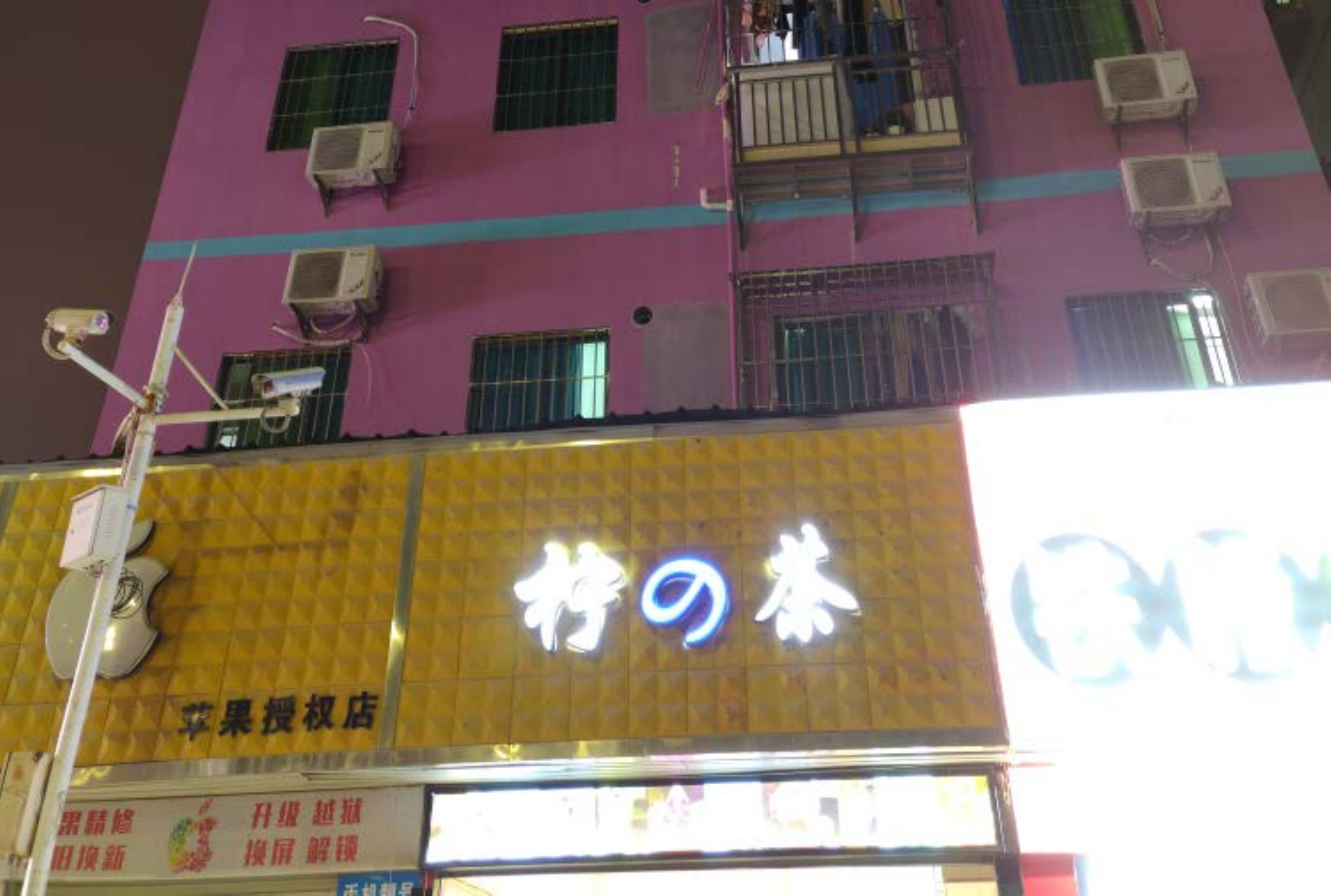} &
        \includegraphics[width=\swthree,angle=0]{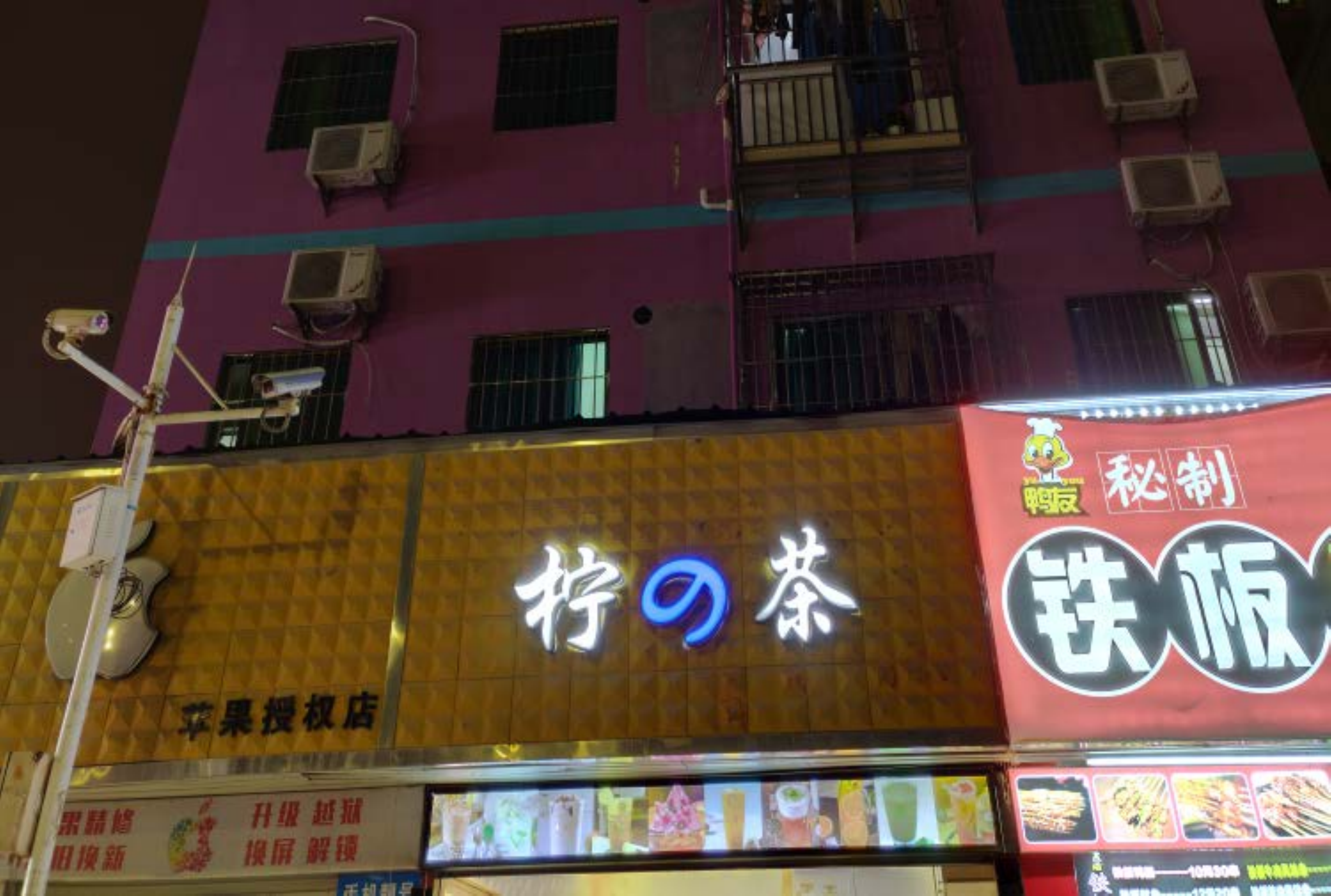} &
        \includegraphics[width=\swthree,angle=0]{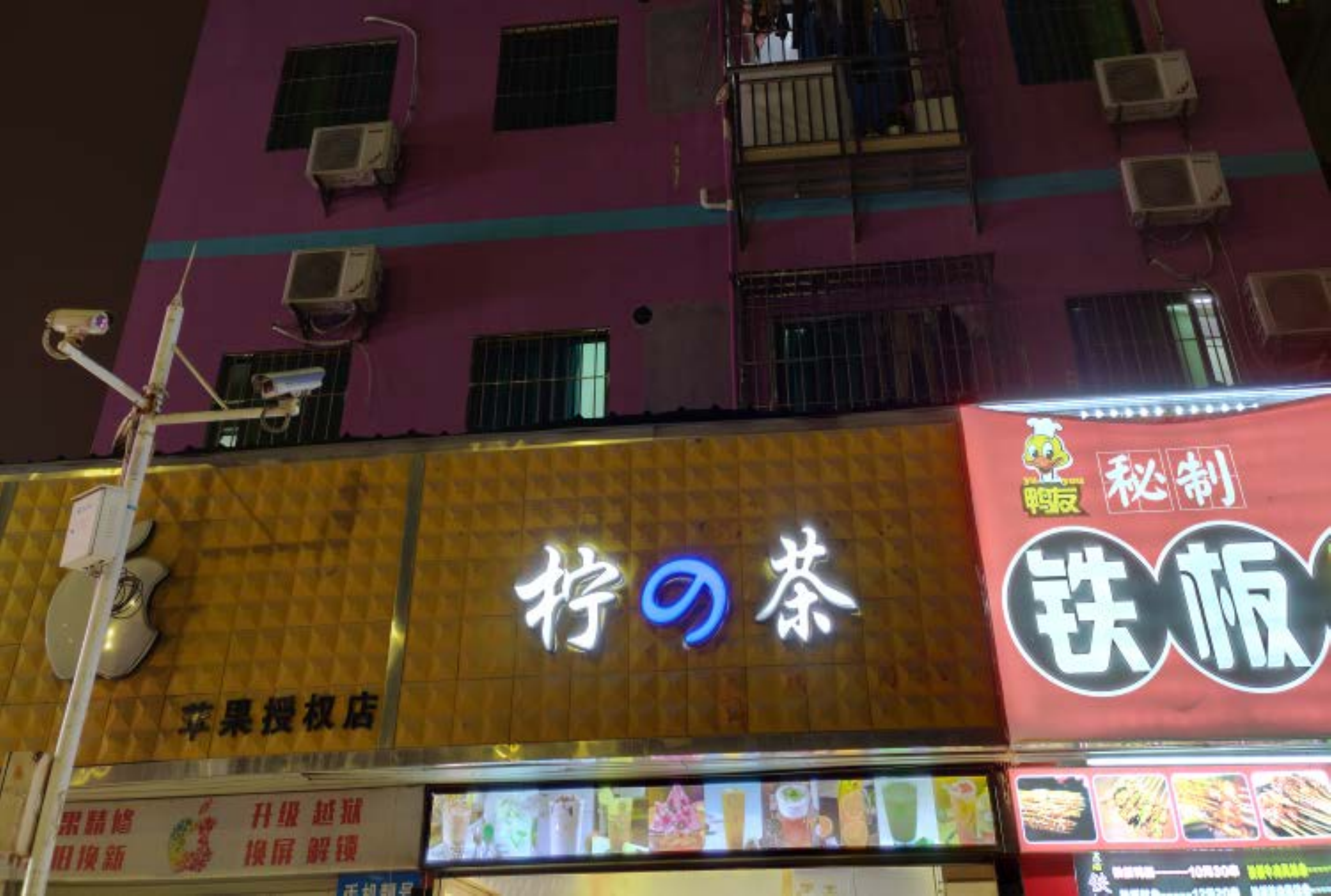} \\
        (a) AE preview $\mathbf{x}$ & (b) Barakat~\cite{barakat2008minimal} & (c) Beek~\cite{van2018improved} \\
        Selected exposure bracketing & $\{\mathbf{z}_3, \mathbf{z}_9\}$ & $\{\mathbf{z}_3, \mathbf{z}_9\}$ \\
        \includegraphics[width=\swthree,angle=0]{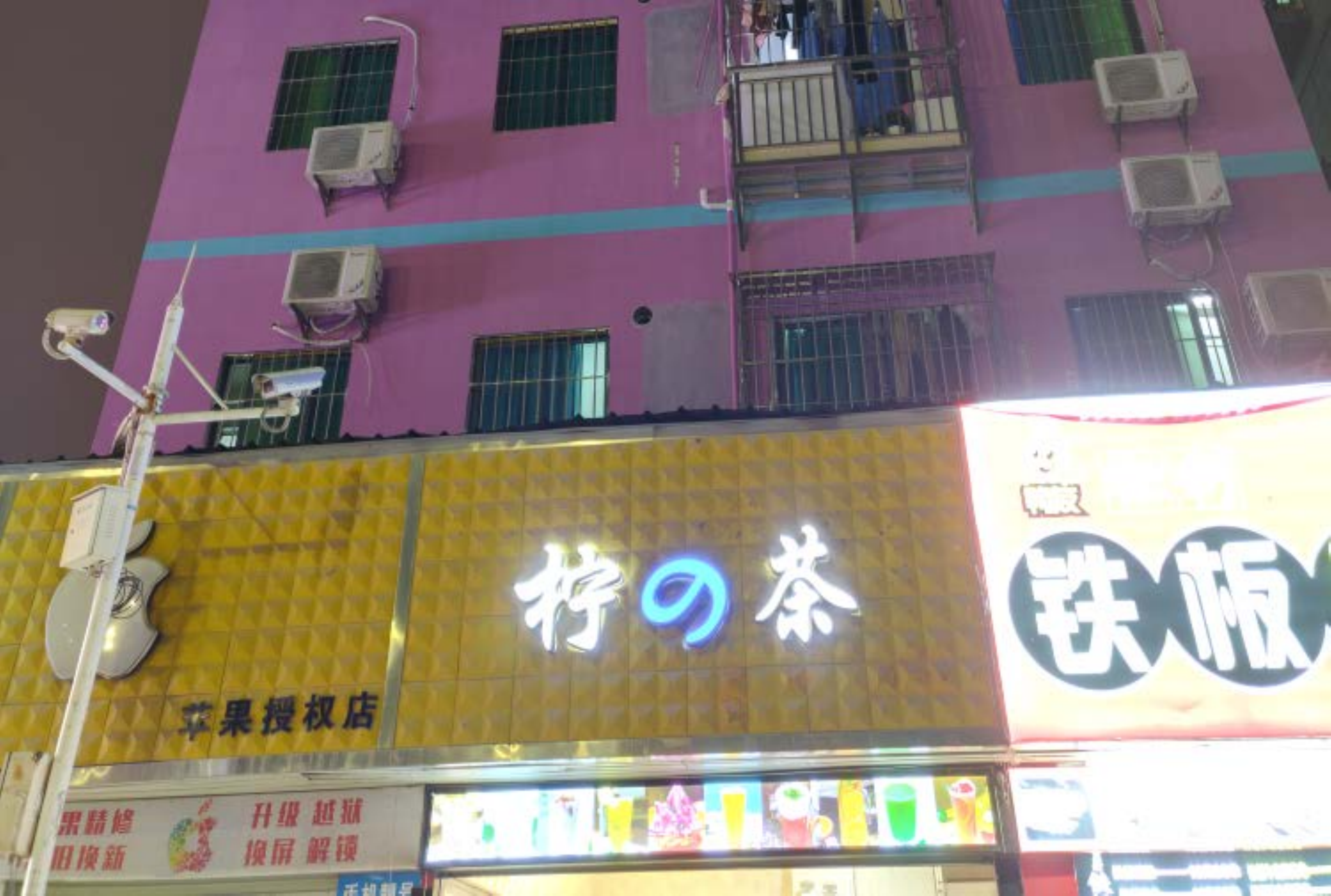} &
        \includegraphics[width=\swthree,angle=0]{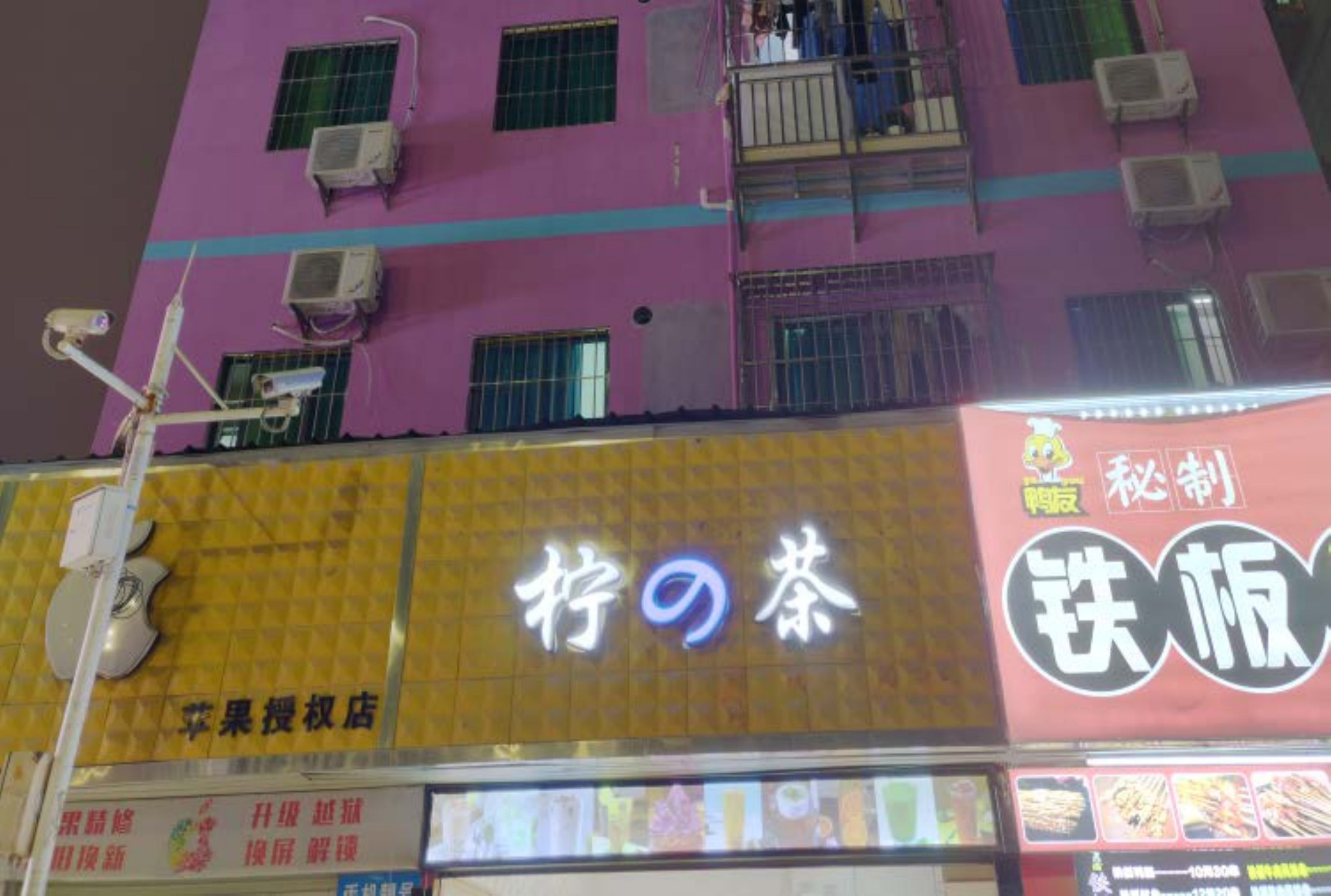} &
        \includegraphics[width=\swthree,angle=0]{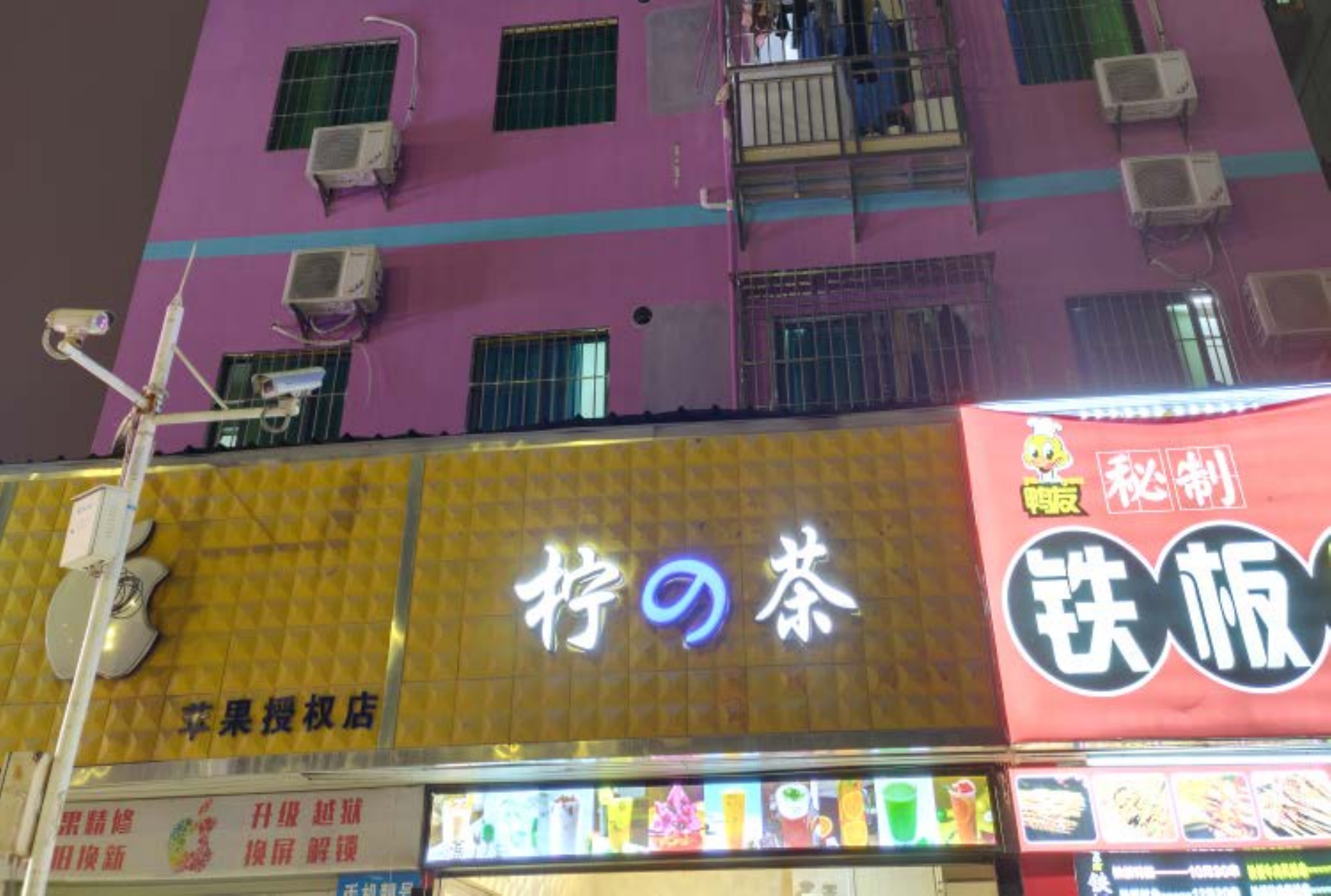}\\
        (d) Pourreza-Shahri~\cite{pourreza2015exposure} & (e) EBSNet+MEFNet & (f) GT \\ 
        $\{\mathbf{z}_0, \mathbf{z}_1, \mathbf{z}_5\}$ & $\{\mathbf{z}_0, \mathbf{z}_2, \mathbf{z}_8\}$ & $\{\mathbf{z}_0, \mathbf{z}_1, \mathbf{z}_2, \mathbf{z}_3, \mathbf{z}_4, \mathbf{z}_5, \mathbf{z}_6, \mathbf{z}_7, \mathbf{z}_8, \mathbf{z}_9\}$\\
        \includegraphics[width=\swthree,angle=0]{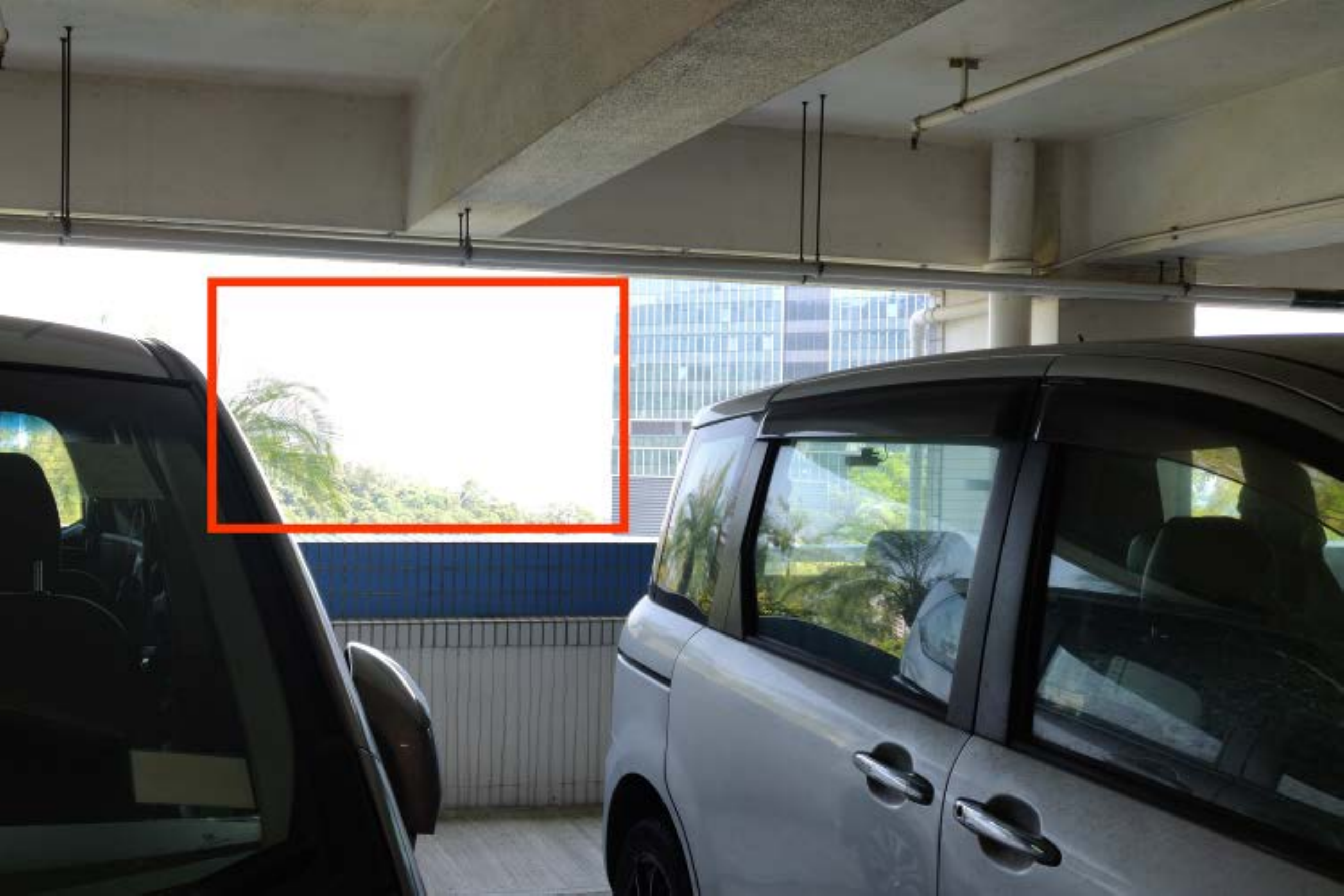} &
        \includegraphics[width=\swthree,angle=0]{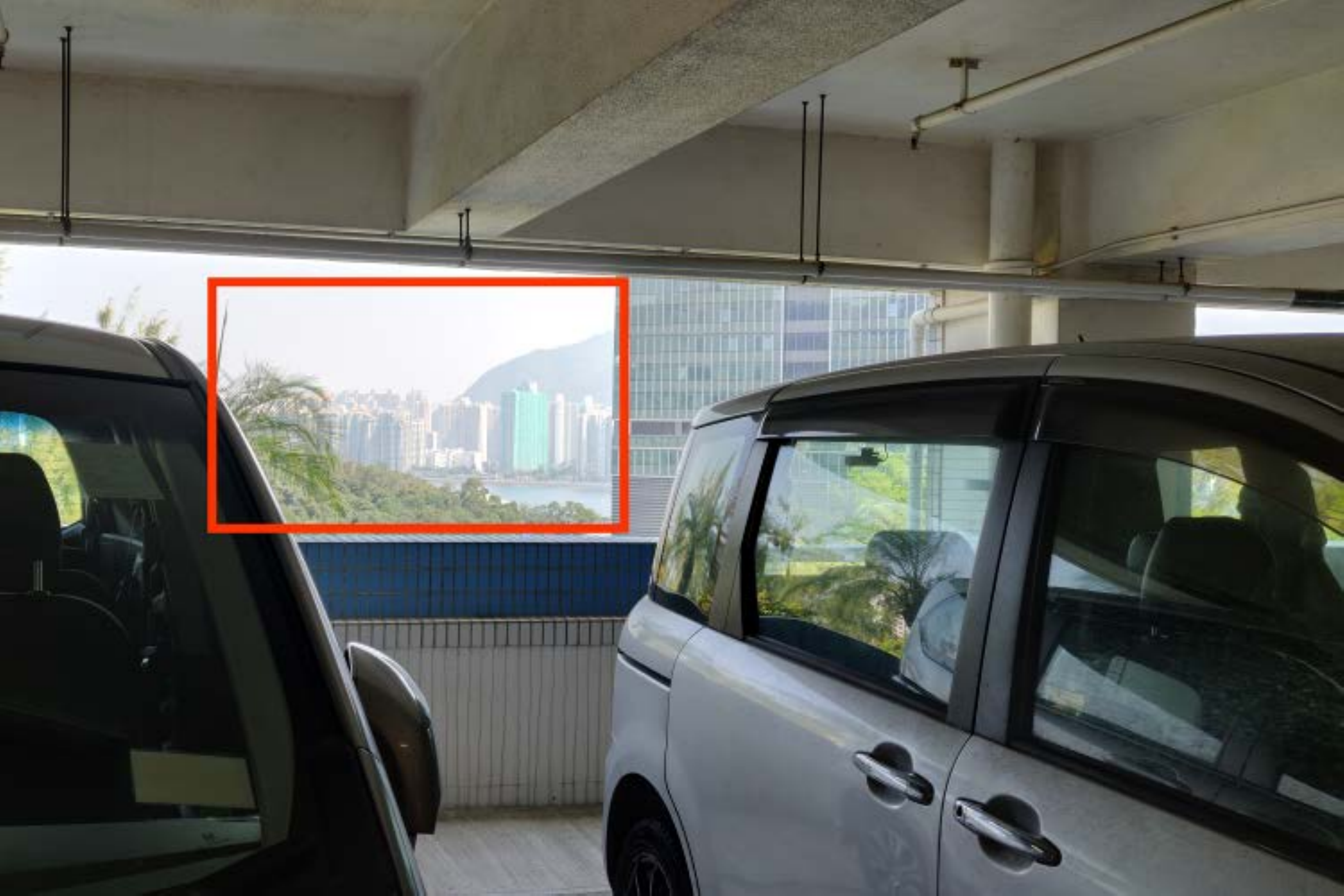} &
        \includegraphics[width=\swthree,angle=0]{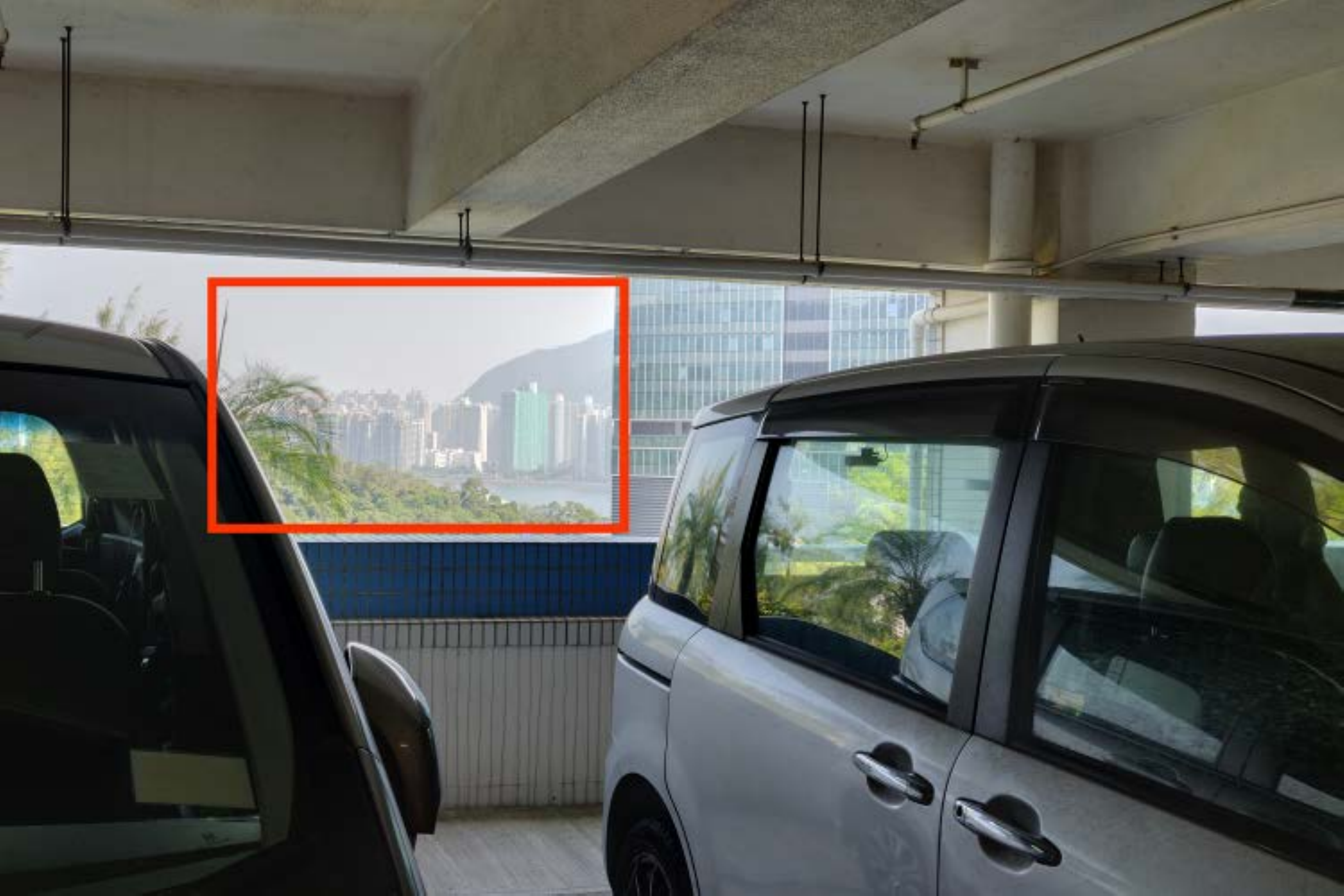} \\
        (g) AE preview $\mathbf{x}$ & (h) Barakat~\cite{barakat2008minimal} & (i) Beek~\cite{van2018improved} \\
        Selected exposure bracketing & $\{\mathbf{z_0}, \mathbf{z_1}, \mathbf{z}_6\}$ & $\{\mathbf{z}_0, \mathbf{z}_1, \mathbf{z}_2, \mathbf{z}_3, \mathbf{z}_6, \mathbf{z}_7, \mathbf{z}_8, \mathbf{z}_9\}$ \\
        \includegraphics[width=\swthree,angle=0]{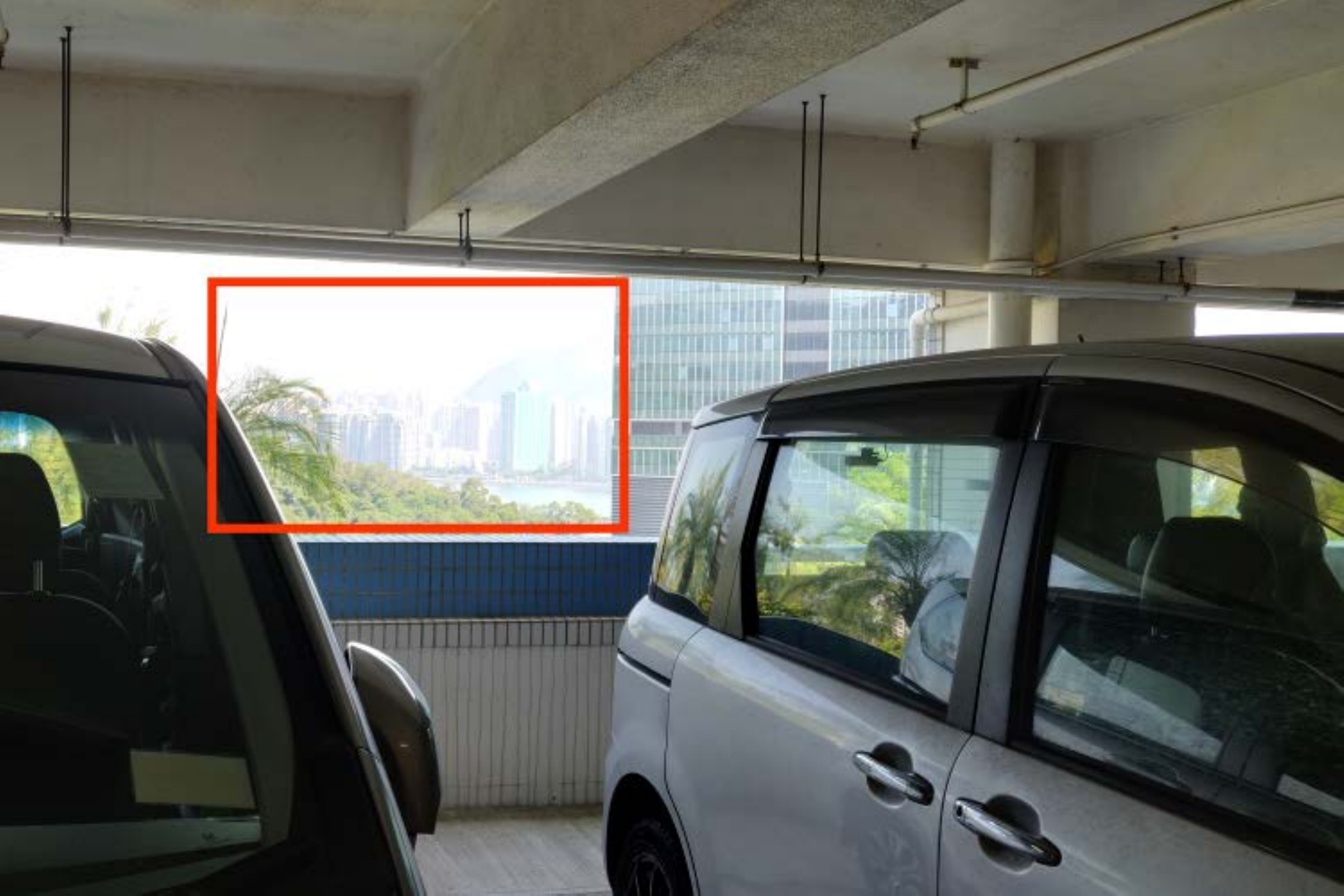} &
        \includegraphics[width=\swthree,angle=0]{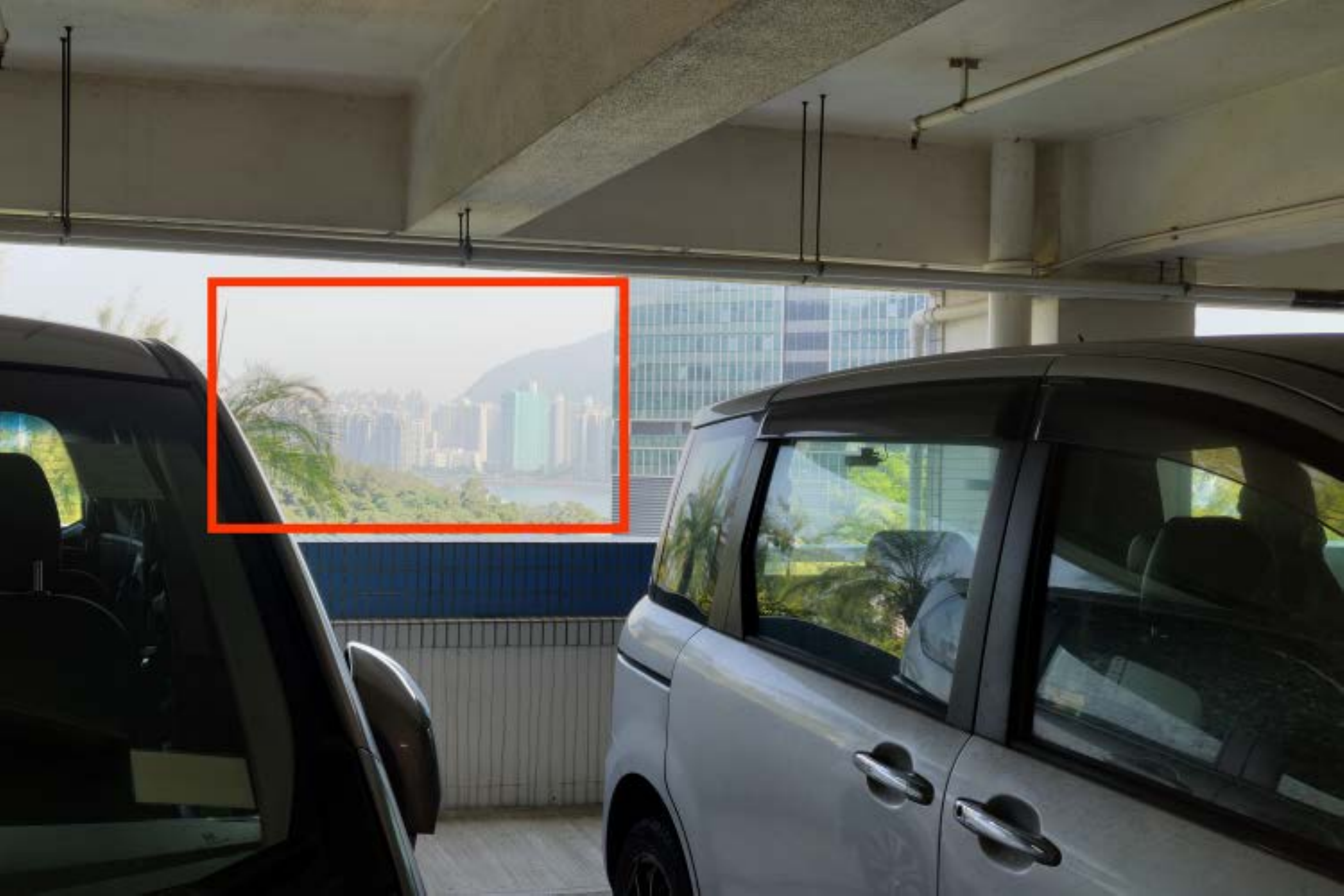} &
        \includegraphics[width=\swthree,angle=0]{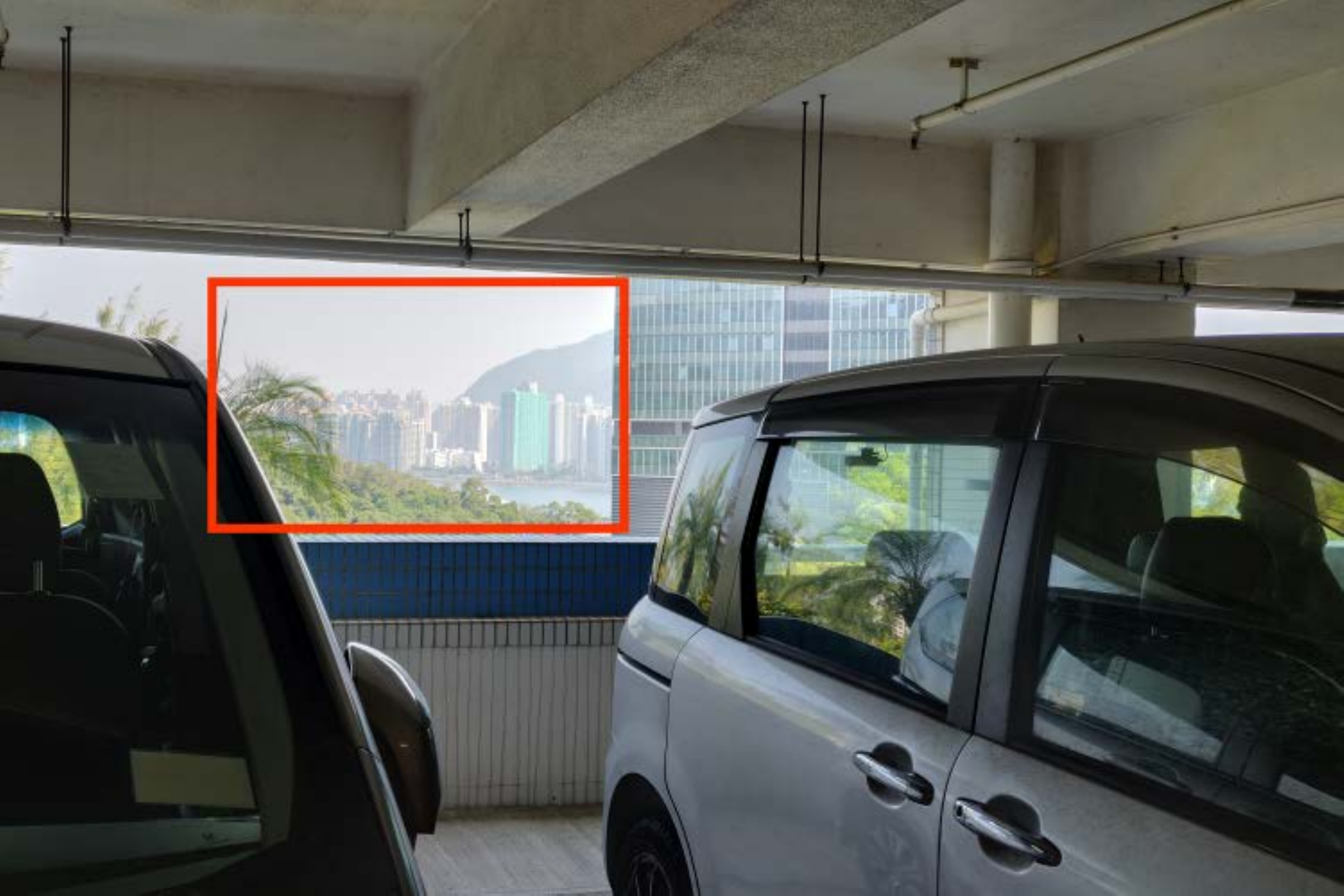} \\
        (j) Pourreza-Shahri~\cite{pourreza2015exposure} & (k) EBSNet+MEFNet & (l) GT \\ 
        $\{\mathbf{z}_0, \mathbf{z}_1, \mathbf{z}_4\}$ & $\{\mathbf{z}_1, \mathbf{z}_3, \mathbf{z}_7\}$ & $\{\mathbf{z}_0, \mathbf{z}_1, \mathbf{z}_2, \mathbf{z}_3, \mathbf{z}_4, \mathbf{z}_5, \mathbf{z}_6, \mathbf{z}_ 7, \mathbf{z}_8, \mathbf{z}_9\}$\\
        \end{tabular}
        \end{center}
        \caption{
        Comparison with the state-of-the-art exposure bracketing selection methods under night and daytime.
        By considering a stream of preview images with different exposures and the camera response function, Barakat et al.~\cite{barakat2008minimal} and Beek et al.~\cite{van2018improved} can recover the saturated regions well, e.g. the billboard in (b)(c) and the building in (h)(i).
        However, their objective is to estimate a minimal bracketing to get a worst-case SNR, they fail to generate a bright enough building in (b) and (c).
        Both Pourreza-Shahri et al.~\cite{pourreza2015exposure} and the proposed method only use one AE preview image to select the exposure bracketing.
        As Pourreza-Shahri et al.~\cite{pourreza2015exposure} fail to consider the semantic information, it cannot recover the saturated regions, e.g. the billboard in (d) and the building in (j).
        With the help of semantic information, the proposed method can generate a better result both in the dark region in (e) and the saturated region in (k).
        Best viewed in color; zoom in for additional details.
        }
        \label{fig:compair}
        \end{figure*}  
        
        \begin{table}[htp]
                \begin{center}
                    \begin{tabular}{|l|c| }
                        \hline
                        Method & PSNR \\
                        \hline\hline\hline
                        Barakat~\cite{barakat2008minimal} & 27.31  \\
                        Pourreza-Shahri~\cite{pourreza2015exposure} & 26.44  \\
                        Beek~\cite{van2018improved} & 27.46  \\
                        EBSNet+EF\cite{mertens2009exposure} & 28.15  \\
                        EBSNet+MEFNnet & 28.41  \\        
                        \hline
                    \end{tabular}
                \end{center}
                \caption{ 
                    Comparison of the state-of-the-art methods and ours in terms of PSNR.
                    Both `EBSNet+EF' and `EBSNet+MEFNet' use the proposed exposure bracketing selection network.
                    The only difference is that `EBSNet+EF' uses exposure fusion proposed by \cite{mertens2009exposure} while `EBSNet+MEFNet' uses the proposed multi-exposure fusion network.
                    It shows that the proposed method can benefit from the joint training of EBSNet and MEFNet.
                    In addition, the proposed method performs better than the existing methods.
                }
                \label{tb:performance}
                \end{table}
        
        \subsection{Comparison with State-of-the-art Methods}
        To validate the effectiveness of our proposed method on exposure bracketing selection, we compare the performance of our method against the state-of-the-art approaches proposed by Barakat et al.~\cite{barakat2008minimal}, Pourreza-Shahri and Kehtarnavaz~\cite{pourreza2015exposure}, and Beek et al.~\cite{van2018improved}.
        
        For a fair comparison, we use the exposure fusion method proposed by \cite{mertens2009exposure} to generate the HDR image with the selected exposure bracketing from the above methods as well as the proposed EBSNet (denoted as EBSNet+EF).
        And our whole framework with both the proposed exposure bracketing selection network as well as the multi-exposure fusion network is denoted as EBSNet+MEFNet.
        Besides, the average number of the selected exposure bracketing of these methods is close to three which is shown in Table~\ref{tb:number}.

\begin{figure*}[htb]\footnotesize
        \begin{center}
        \renewcommand{\tabcolsep}{1pt}
        \begin{tabular}{cccc}
            \includegraphics[width=\swfour,angle=0]{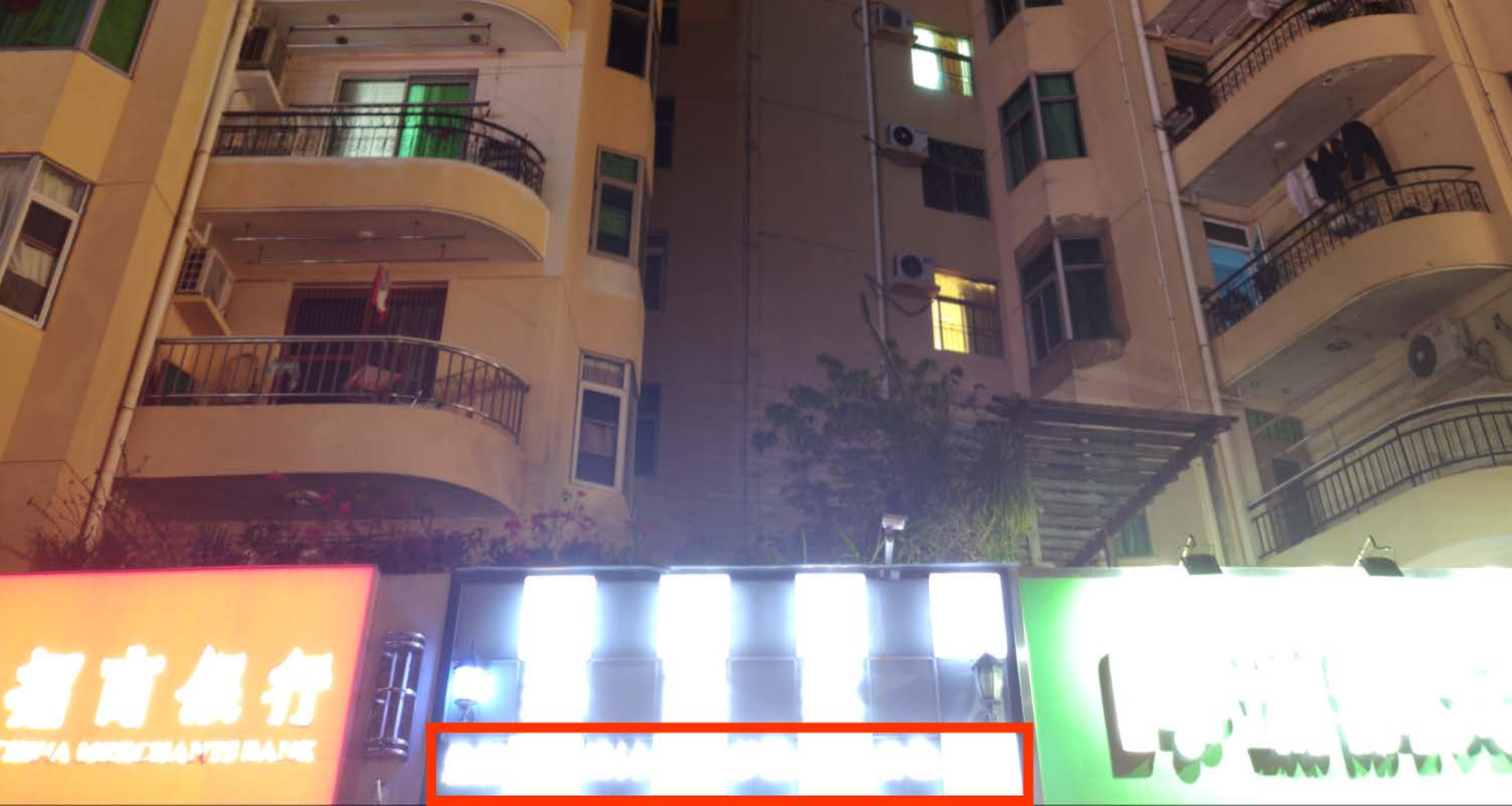} &
            \includegraphics[width=\swfour,angle=0]{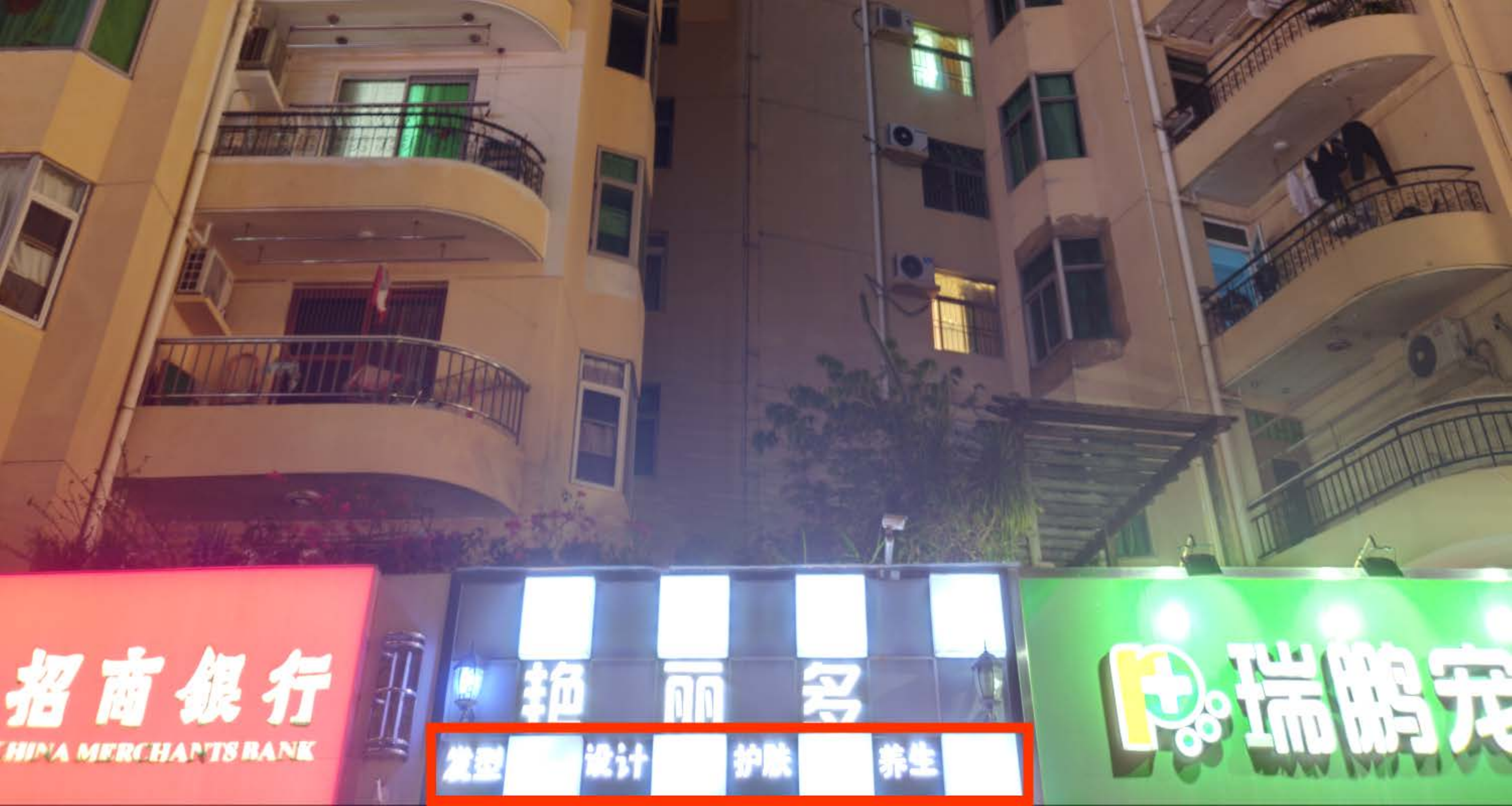} &
            \includegraphics[width=\swfour,angle=0]{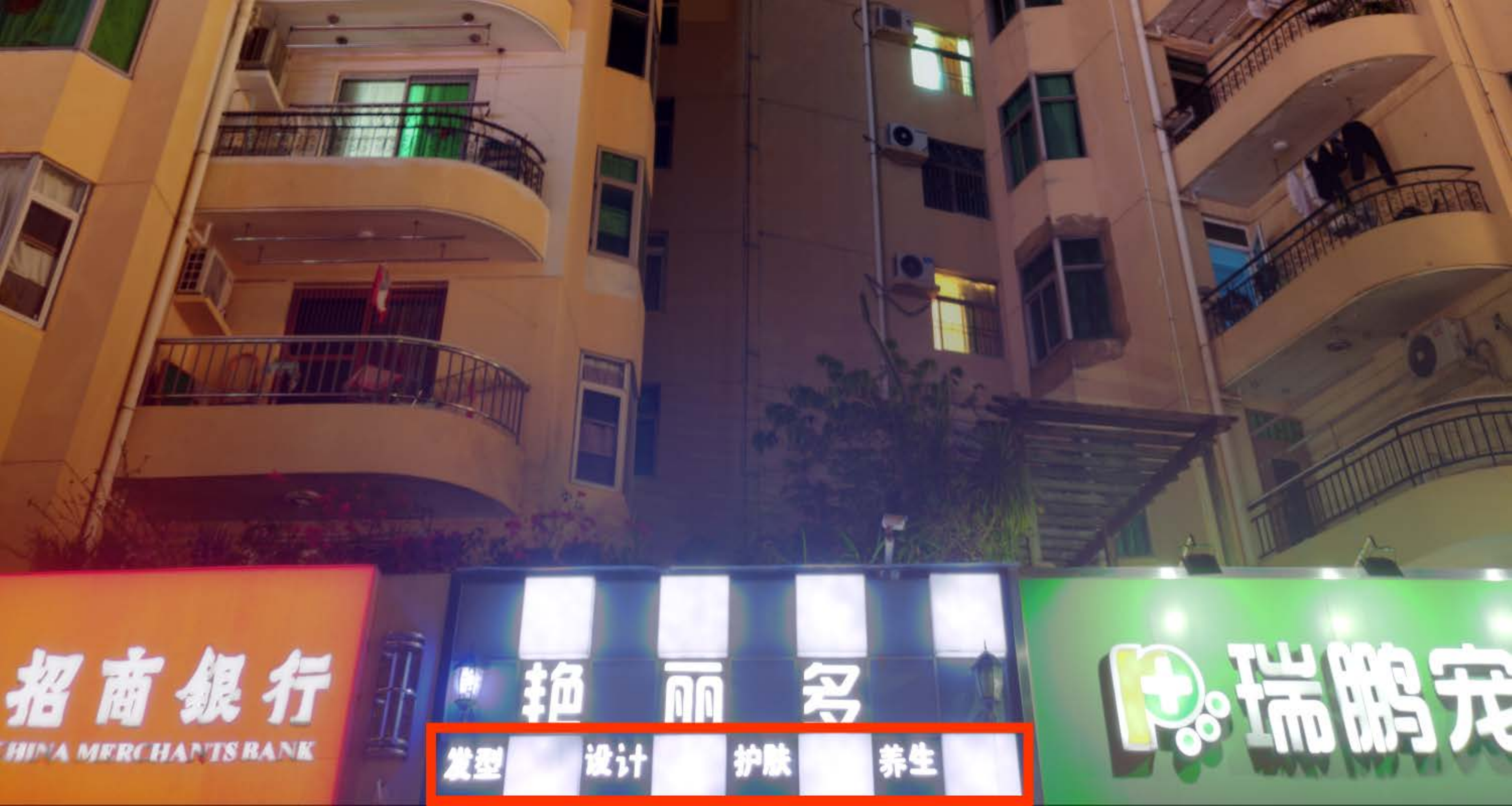} &
            \includegraphics[width=\swfour,angle=0]{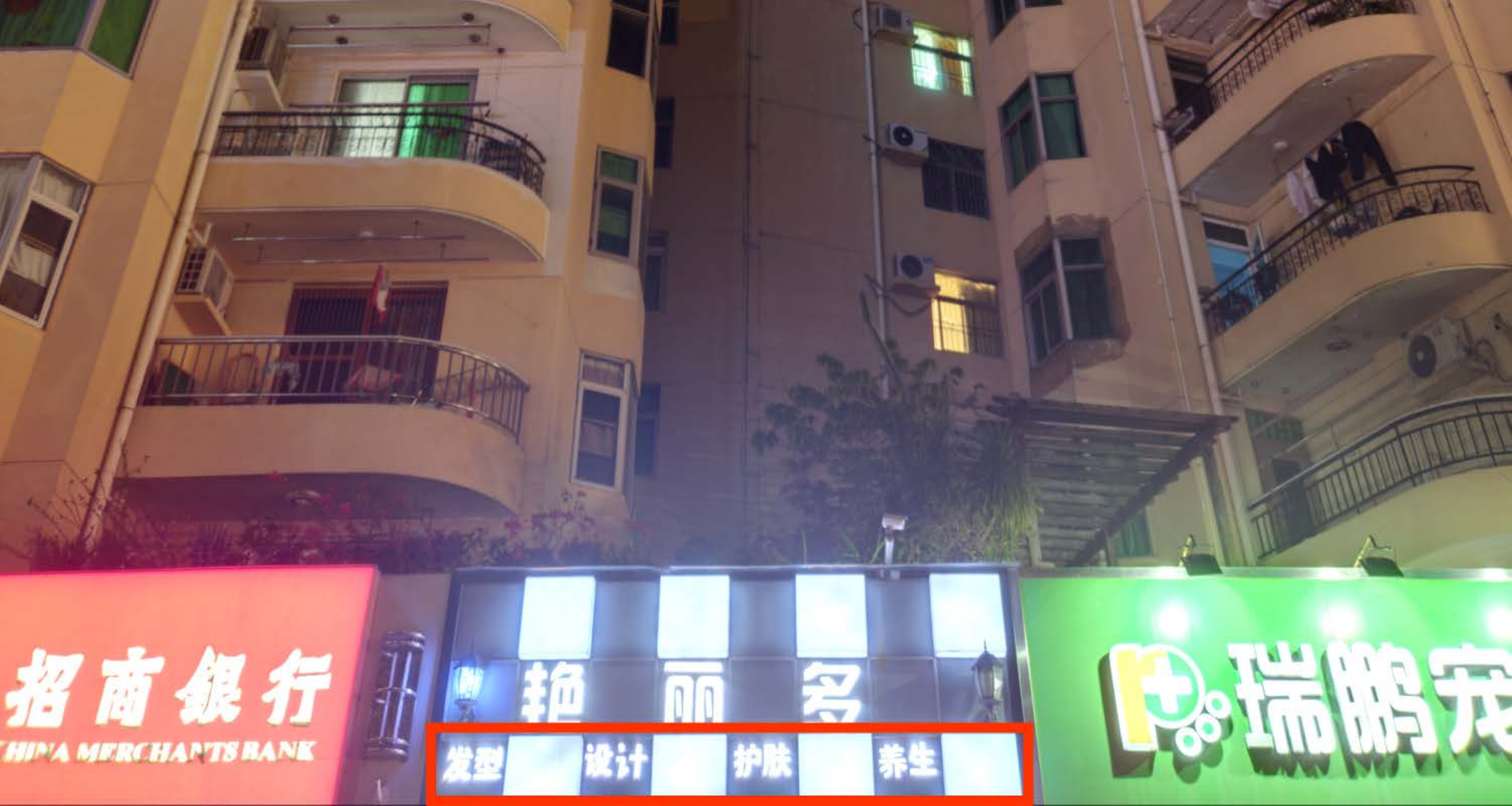} \\
            \includegraphics[width=\swfour,angle=0]{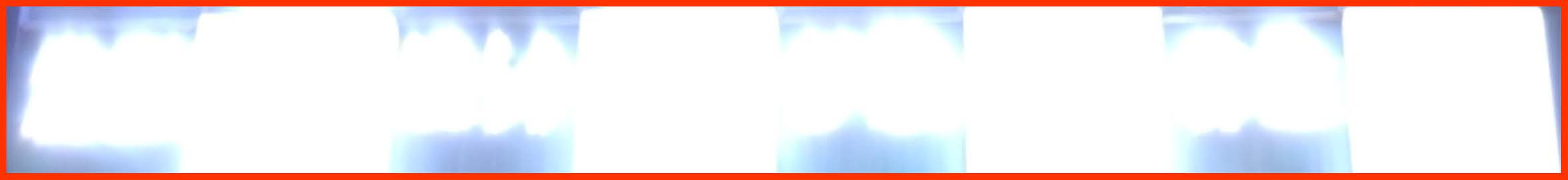} &
            \includegraphics[width=\swfour,angle=0]{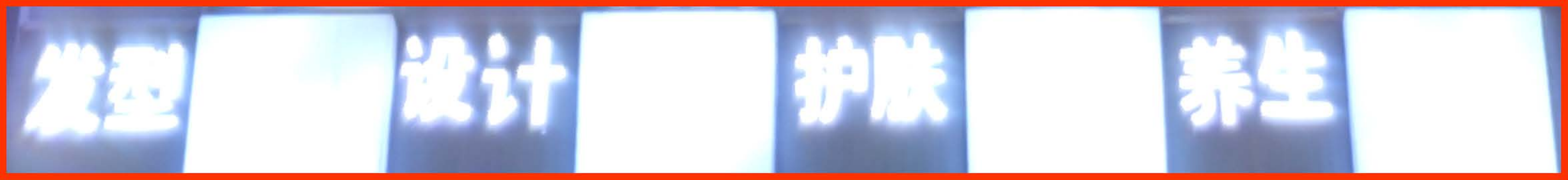} &
            \includegraphics[width=\swfour,angle=0]{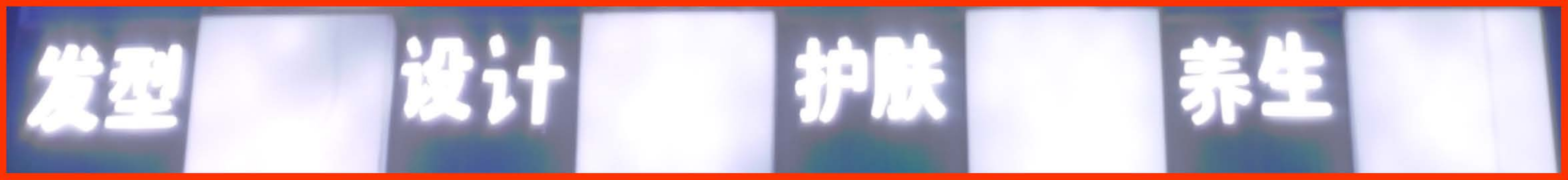} &
            \includegraphics[width=\swfour,angle=0]{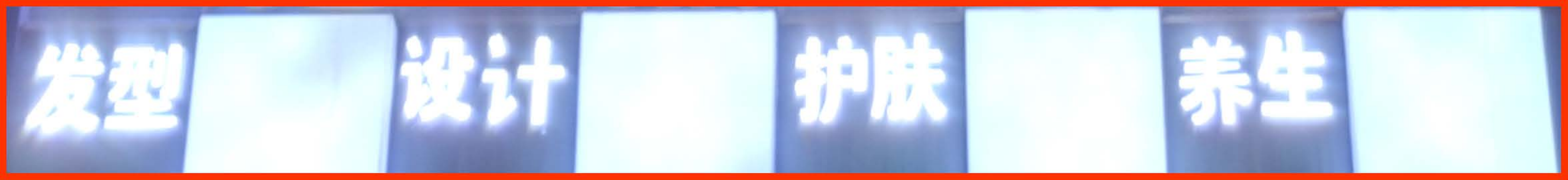} \\
            (a) AE preview $\mathbf{x}$ & (b)  EBSNet+EF~\cite{mertens2009exposure} & (c) EBSNet+MEFNet & (d) GT \\
        \end{tabular}
        \end{center}
        \vspace{-2mm}
        \caption{
            Effectiveness of joint training.
            (a) is the AE preview image, (b) and (c) use \cite{mertens2009exposure} and the proposed MEFNet to generate HDR image with the exposure bracketing selected by the proposed EBSNet respectively, (d) is the ground truth.
            With jointly training EBSNet with MEFNet, MEFNet can suppress the saturated regions better than  Mertens et al.~\cite{mertens2009exposure} especially for the characters in the billboard since our objective is to make the fusion result of MEFNet as close as to the ground truth.
            Best viewed in color; zoom in for additional details.
        }
        \label{fig:ex}
        \end{figure*}

        According to Table~\ref{tb:performance}, EBSNet+EF performs comparable relative to the existing methods.
        And EBSNet+MENFet can further improve the performance through joint training which can also be seen from Figure~\ref{fig:ex}. 
        As shown in Figure~\ref{fig:compair}, both the billboard at night and the building outside in the daytime are saturated in the AE preview $\mathbf{x}$.
        The methods proposed by Barakat et al.~\cite{barakat2008minimal} and Beek et al.~\cite{van2018improved} are based on the estimation of the dynamic range from a stream of preview images with different exposures.
        In this way, they can consider the histogram of the whole scene irradiance for selection and the saturated regions can be recovered well.
        As their objective is to estimate a minimal bracketing to achieve a worst-case SNR, they tend to choose the short exposure.
        As a result, the generated dark regions, e.g. the pink building, seem not as bright as the ground truth.
        Since Pourreza-Shahri et al.~\cite{pourreza2015exposure} only considers the illumination of the preview and does not know the brightness of the saturated region, it fails to recover these regions.
        By considering semantic information through training, the proposed method can better recover the saturated regions (the billboard in (e) and the building in (k)) and keep the brightness of dark regions (the pink building in (e)).
        
        \begin{figure*}[htb]\footnotesize
        \begin{center}
        \renewcommand{\tabcolsep}{1pt}
        \begin{tabular}{cccc}
            \includegraphics[width=\swfour,angle=0]{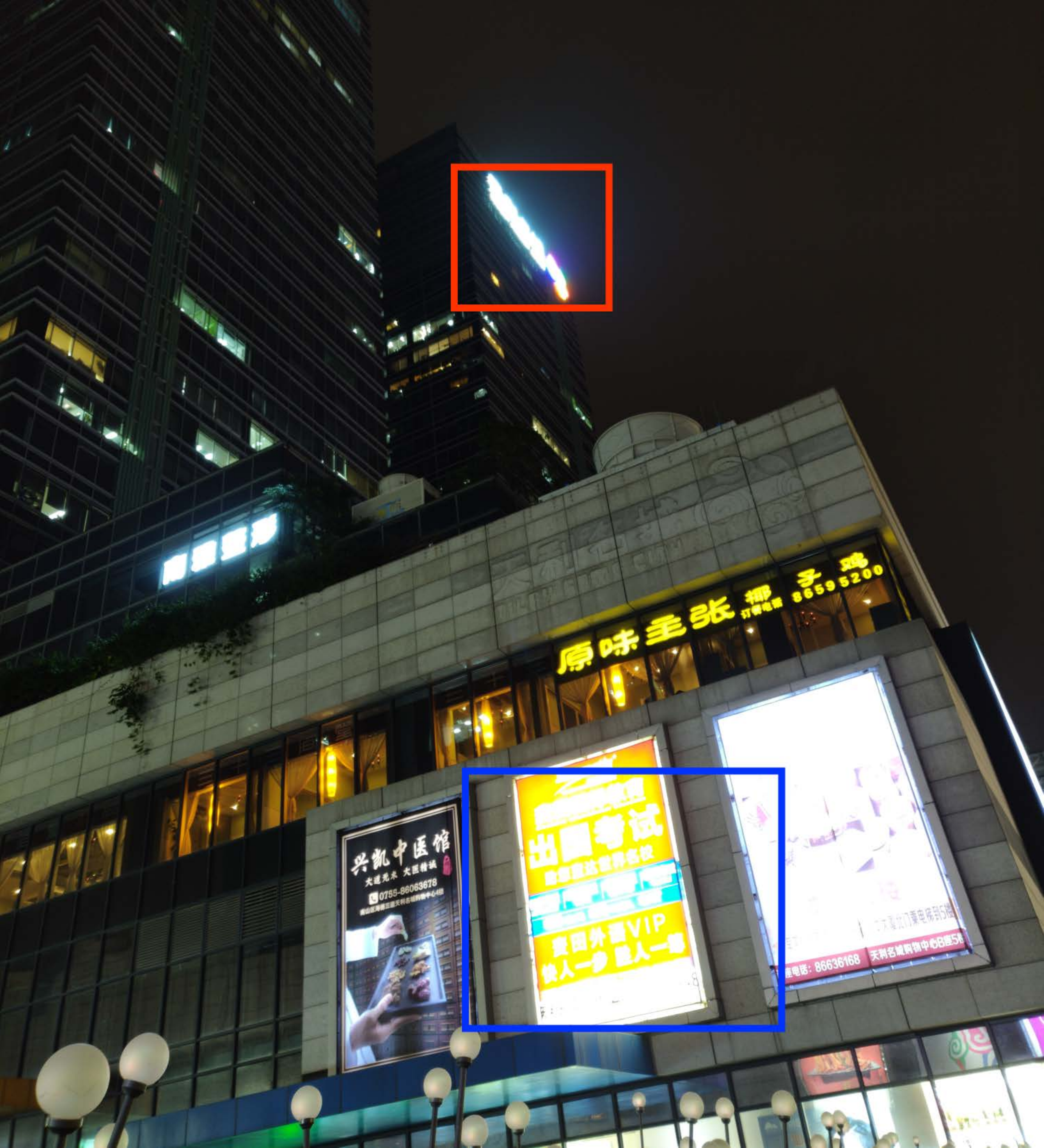} &
            \includegraphics[width=\swfour,angle=0]{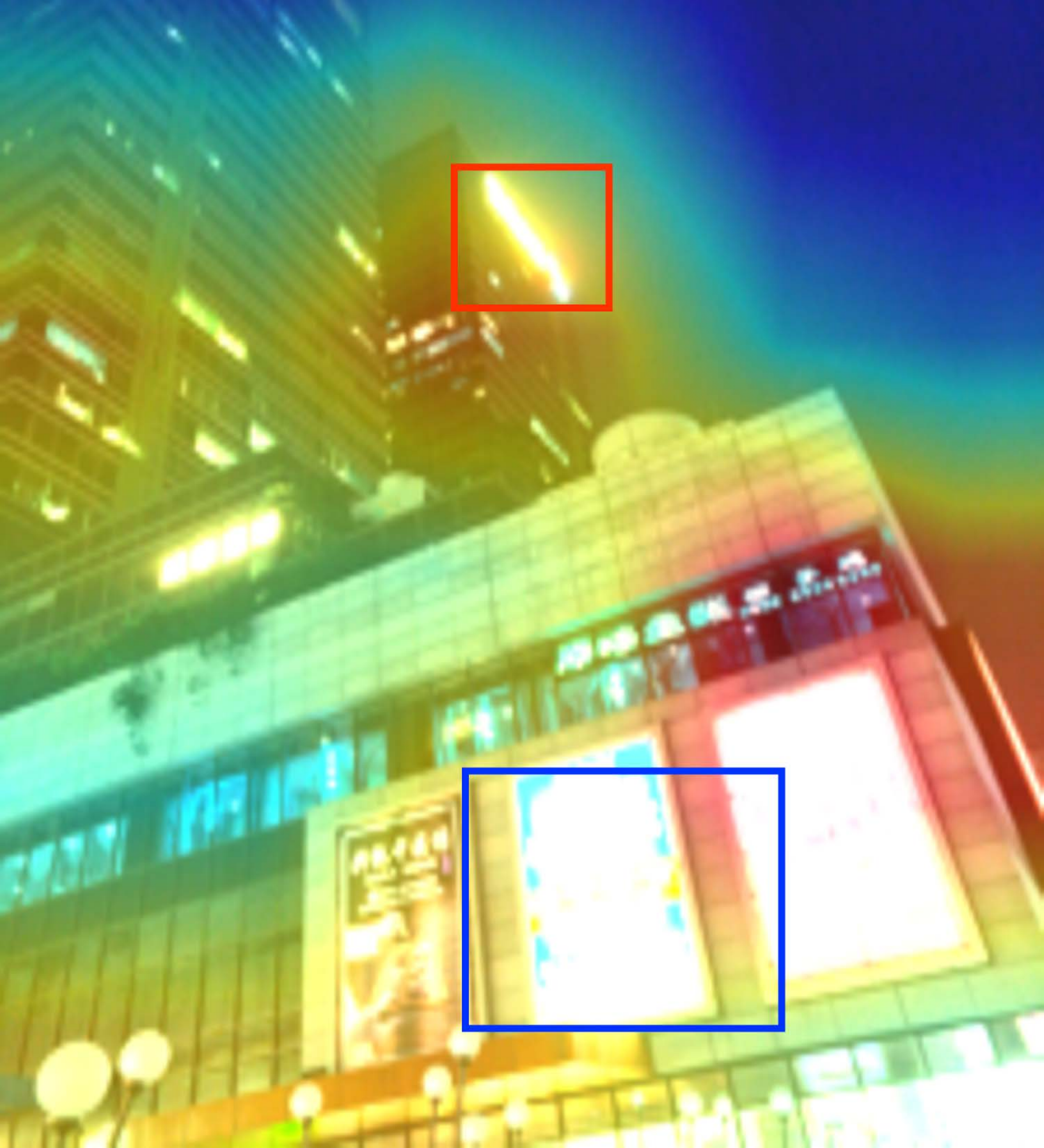} &
            \includegraphics[width=\swfour,angle=0]{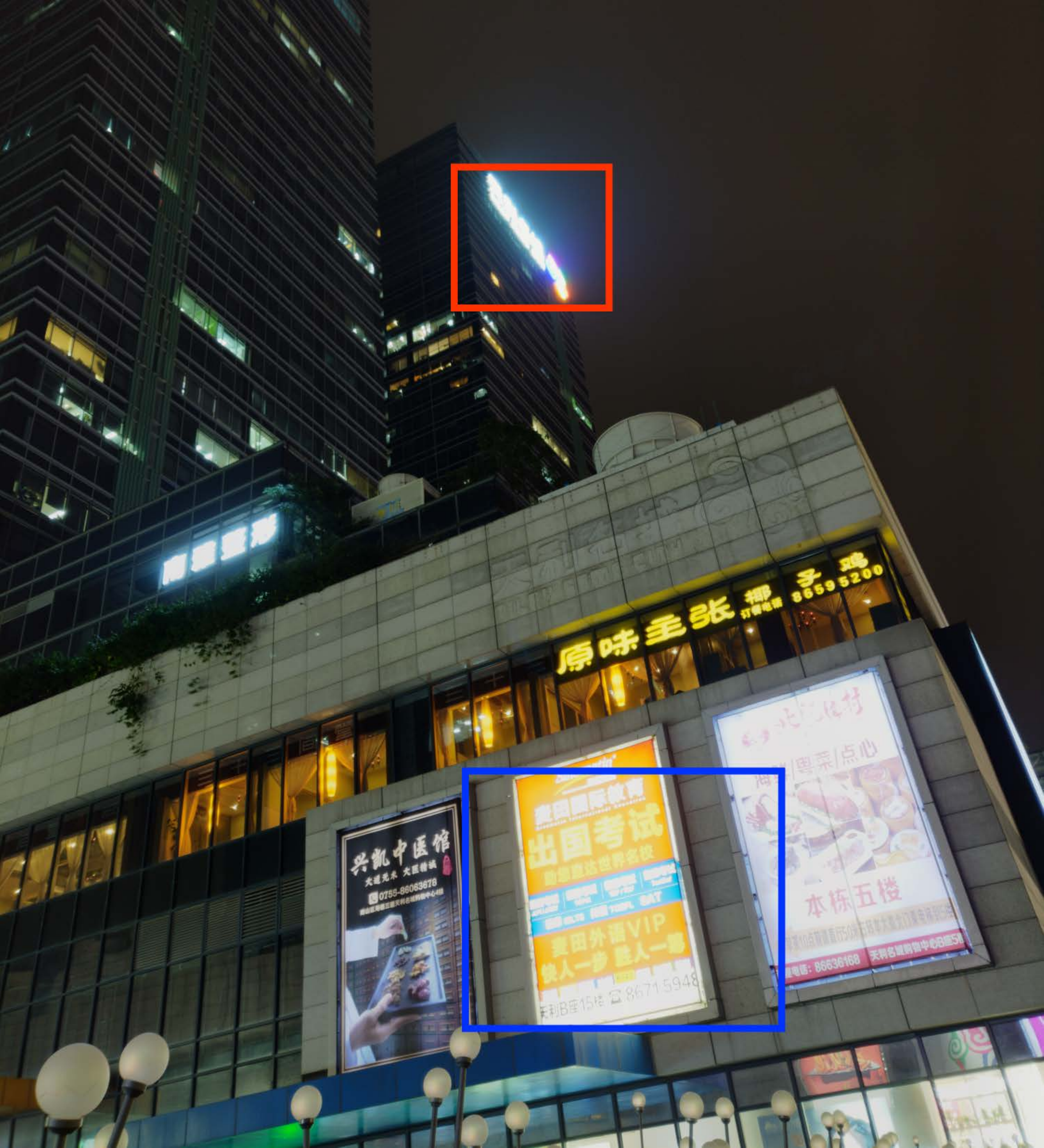} &
            \includegraphics[width=\swfour,angle=0]{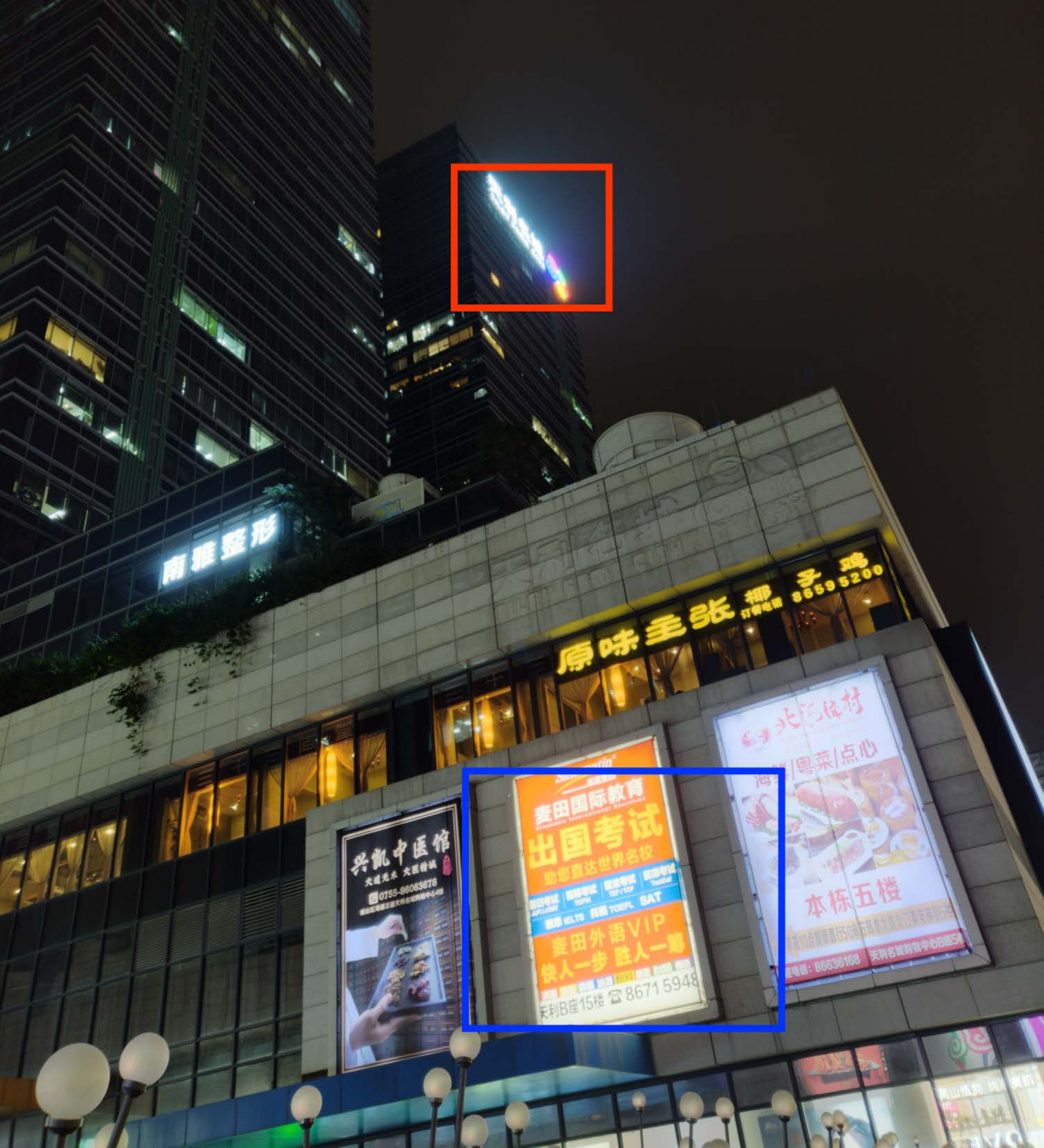} \\
            \includegraphics[width=\swfour,angle=0]{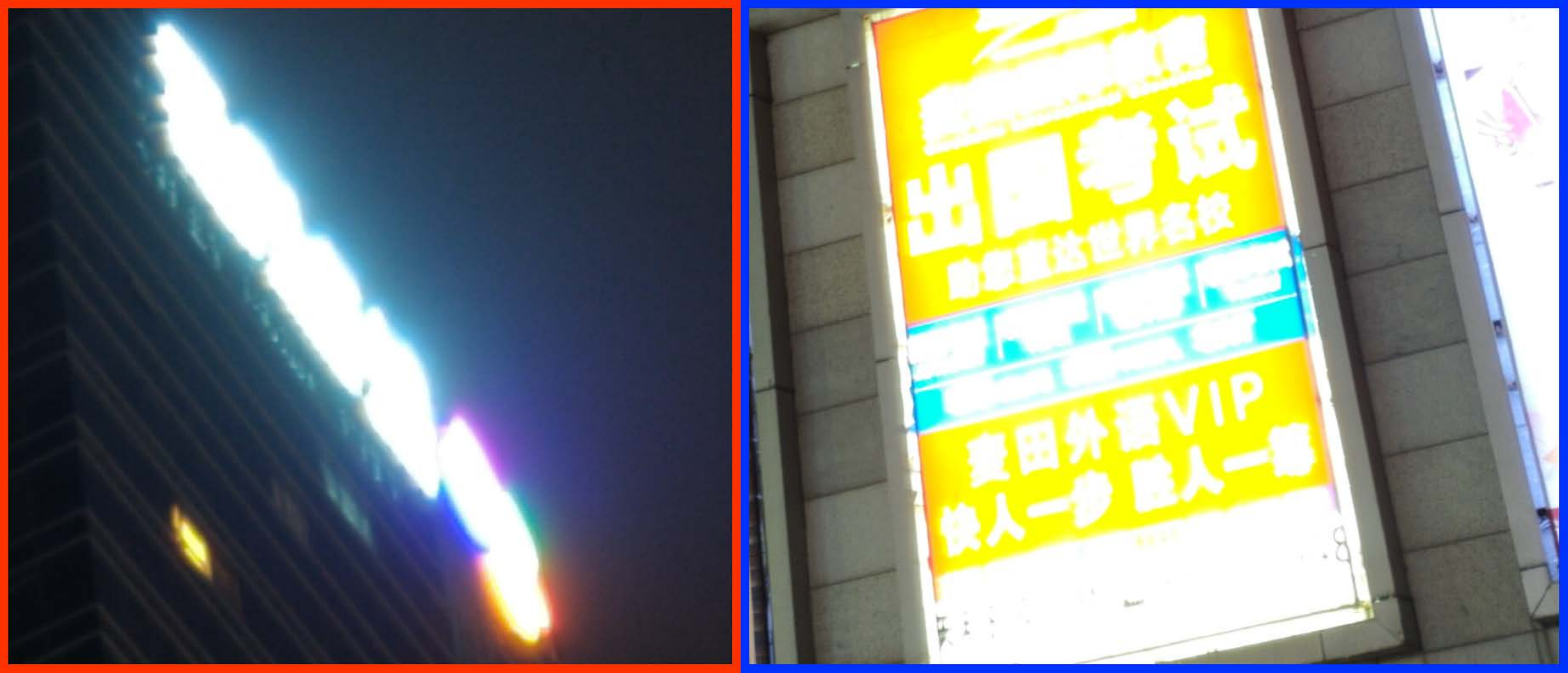} &
            \includegraphics[width=\swfour,angle=0]{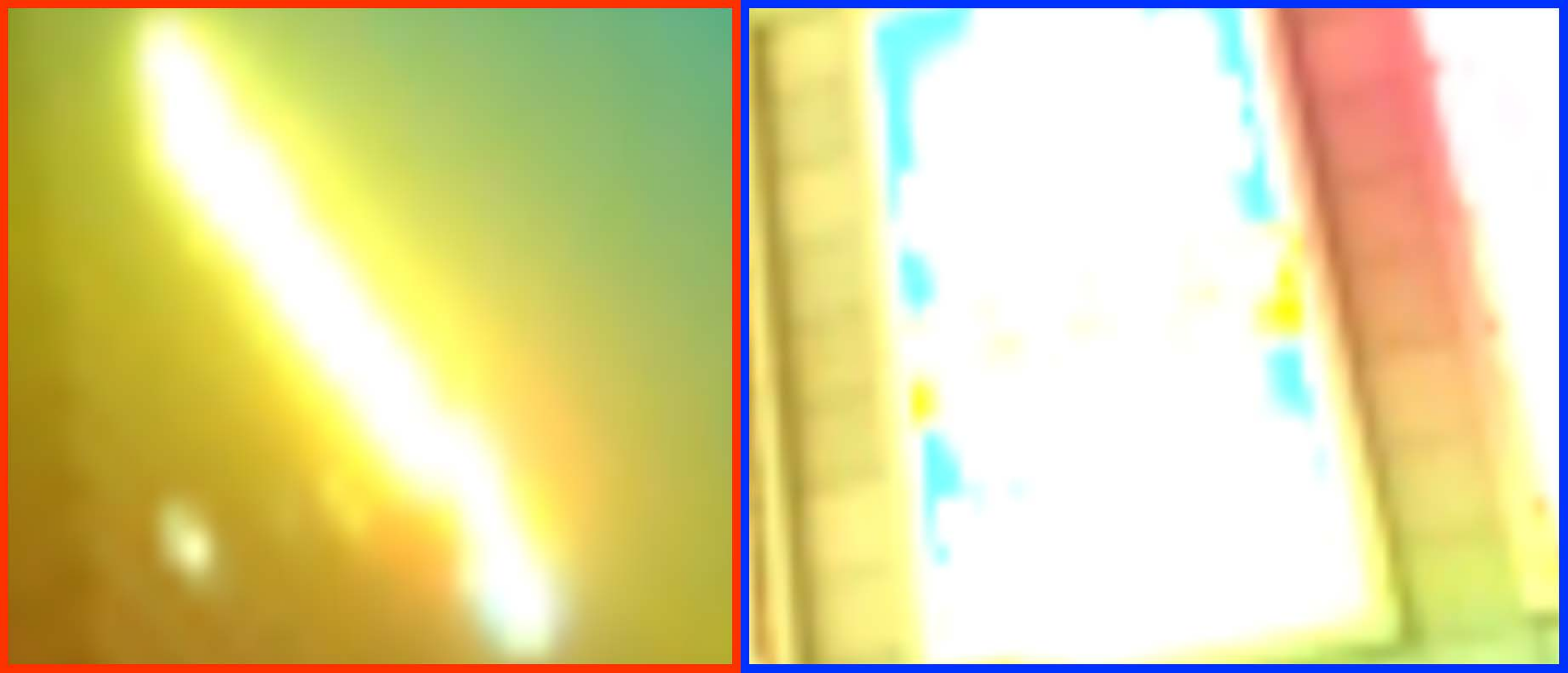} &
            \includegraphics[width=\swfour,angle=0]{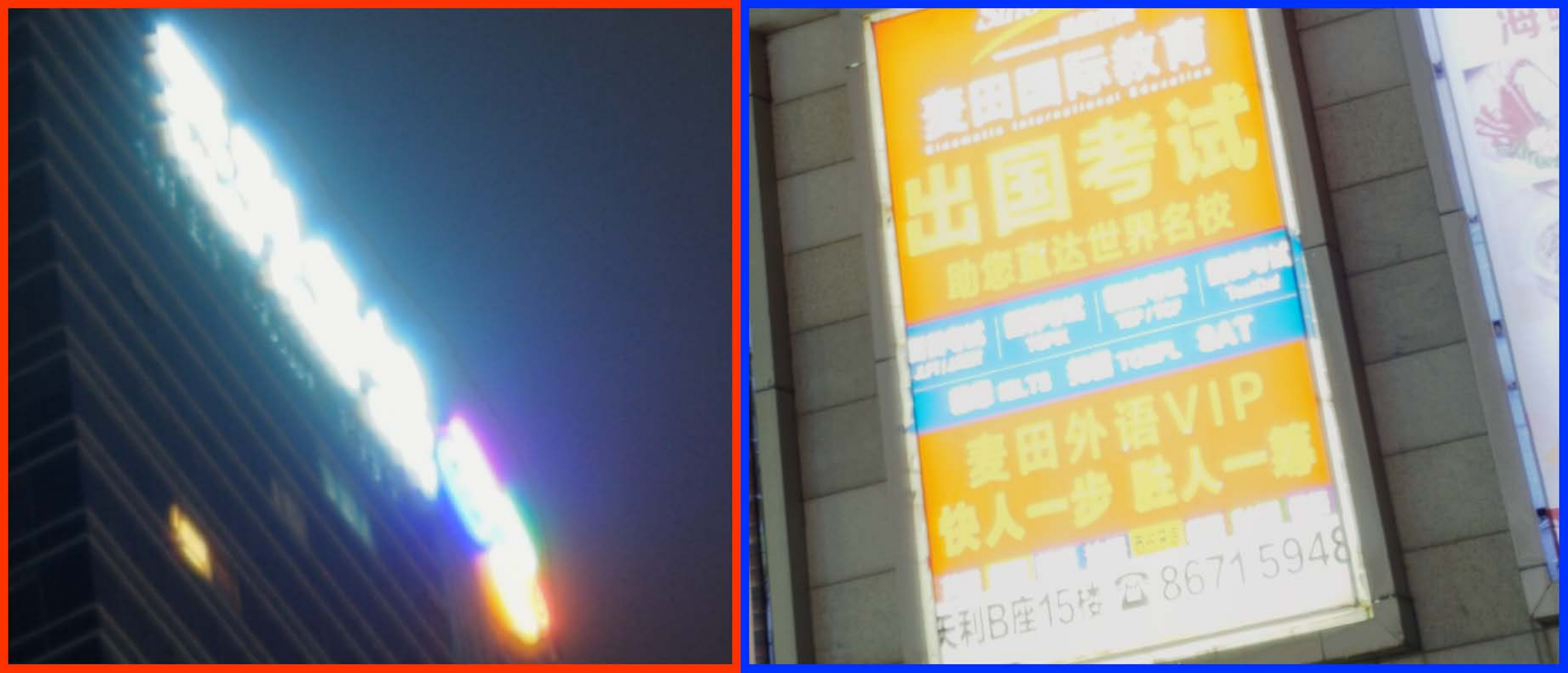} &
            \includegraphics[width=\swfour,angle=0]{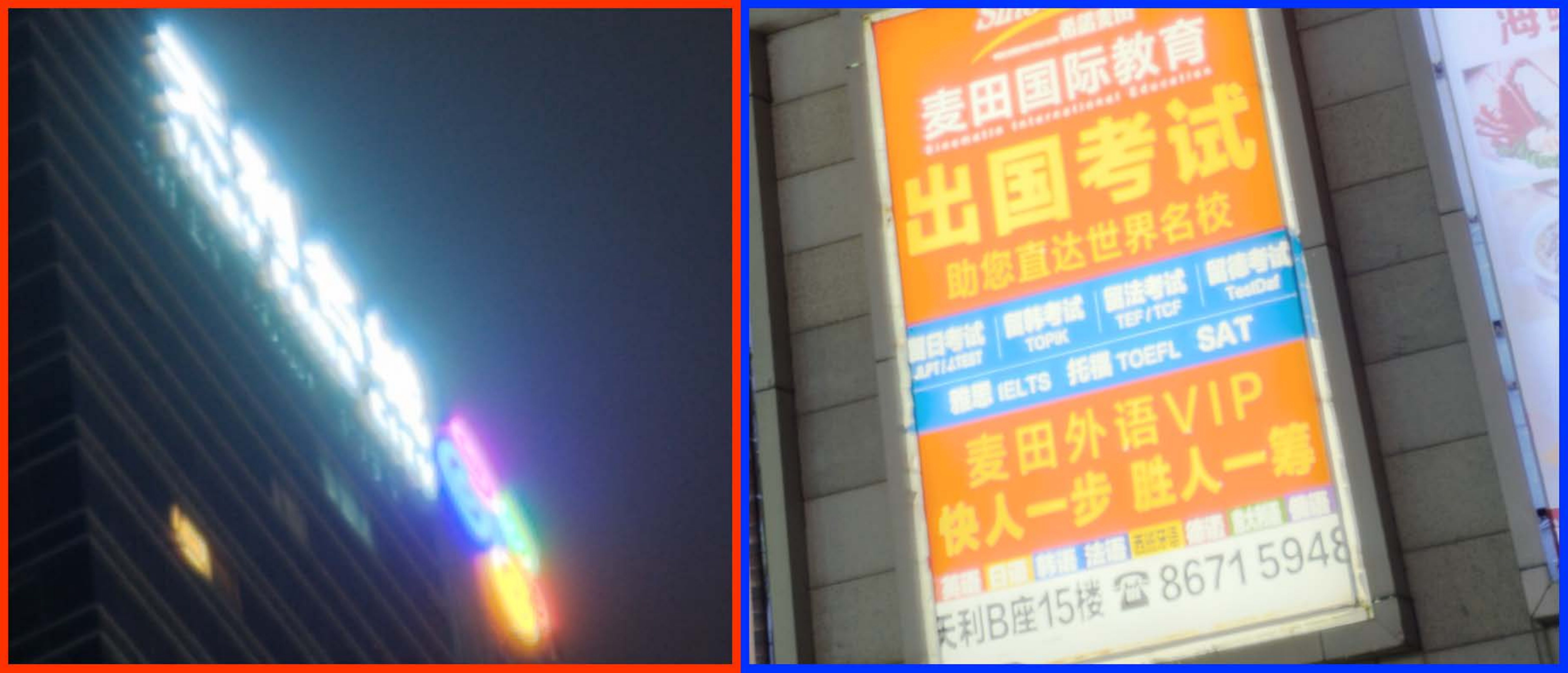} \\
            (a) AE preview $\mathbf{x}$ & (b) Feature Map & (c) Illumination & (d) Illumination+Semantic \\
            Selected exposure bracketing & & $\{\mathbf{z}_0, \mathbf{z}_1, \mathbf{z}_4\}$ & $\{\mathbf{z}_1, \mathbf{z}_4, \mathbf{z}_7\}$ \\
        \end{tabular}
        \end{center}
        \vspace{-2mm}
        \caption{
        Effectiveness of the semantic branch.
        (a) is the AE preview image, (b) is the feature map from the semantic branch of the proposed EBSNet, (c) and (d) are the generated HDR images from EBSNet without and with the semantic branch.
        The feature map from the semantic branch pays more attention to the saturated regions according to (b).
        As a result, (d) decides to select shorter exposures $\{\mathbf{z}_1, \mathbf{z}_4, \mathbf{z}_7\}$ and can recover the saturated regions better than (c).
        Best viewed in color; zoom in for additional details.
        }
       \label{fig:feature}
       \end{figure*}

        \begin{table}
        \begin{center}
        \begin{tabular}{|l|c|c|}
        \hline
        Method & PSNR \\
        \hline\hline\hline
        Semantic & 27.15  \\
        Illumination  & 27.40  \\
        EBSNet+MEFNet & 28.41  \\        
        \hline
        \end{tabular}
        \end{center}
        \caption{ 
        Effectiveness of semantic information according to PSNR.
        The proposed method contains both the semantic branch as well as the illumination branch.
        Even though only the illumination branch performs better than only the semantic branch, the proposed method can achieve better results by considering the semantic information.
        }
        \label{tb:sem_illu}
        \end{table}
        
        \begin{table}
        \begin{center}
        \begin{tabular}{|l|c|c|c|c|}
            \hline
            $K$ & 1 & 2  & 3 & 10 \\
            \hline\hline\hline
            PSNR & 26.06 & 27.53 & 28.41 & 30.14 \\
            \hline
            time/ms & 218.43 & 301.21 & 352.27 & 811.25 \\
            \hline
            \end{tabular}
            \end{center}
            \vspace{-2mm}
            \caption{ 
            Performance and efficiency of different numbers of images $K$ in the selected exposure bracketing in terms of PSNR and running time.
            The running time is about MEFNet in a mainstream Android device.
            The input size is $3000 \times 4000$.
            The proposed method will generate better results with a larger number of images.
            Though the performance is higher when $K=10$, it costs a lot of time while generating the HDR image.
            We choose to use $K=3$ in the proposed method.
            }
        \label{tb:K}
        \end{table}

        \subsection{Ablation study}
        \paragraph{Illumination Distribution v.s. Semantic Information}
        In order to validate the effectiveness of semantic information in EBSNet, we compare the proposed framework with two variant networks: removing the semantic branch and removing the illumination branch.
        Table~\ref{tb:sem_illu} shows that the proposed method performs worse if either of the part is removed.
        The feature map in Figure~\ref{fig:feature}~(b) shows that the network trained with semantic branch pays more attention to the  lights as well as billboards and Figure~\ref{fig:feature}~(d) can suppress the saturated regions better than Figure~\ref{fig:feature}~(c) which is only besed on illumination information.

        \paragraph{Determination of $K$}
        \label{Sec:K}
        Even though Barakat et al.~\cite{barakat2008minimal} have shown that three images can capture the luminance of a scene adequately in most cases, we also take some experiments to compare the different number of images $K$ in the selected exposure bracketing.
        The results are shown in Table~\ref{tb:K}.
        When $K$ equals to 1, it is a single image enhancement task like HDRNet~\cite{gharbi2017deep}.
        It shows that the proposed framework will generate a more accurate HDR image with more LDR images as the input of MEFNet.
        When $K$ equals to 10, the proposed MEFNet considers all the candidate exposures and generates the HDR images the most similar to those from the \cite{mertens2009exposure}.
        Though the performance is higher when $K=10$, it is time-consuming.
        As shown in the table, the computational cost of MEFNet increases with more selected exposures, especially when we fuse the $K$ images with full-resolution ($3000 \times 4000$ of the proposal dataset).
        For the trade-off of performance and efficiency, we choose to use $K=3$ in the proposed method.
        Besides, the running time of EBSNet is about 14ms with any $K$ since it only influences the output size of the last fully-connected layer of EBSNet ($C_{10}^K$).
        We test our model in a mainstream Android smartphone.
       
       \section{Conclusion}
        In this paper, we propose a neural network EBSNet to select exposure bracketing by extracting both the illumination as well as the semantic information from only one single AE preview image.
        It is free from knowing any additional information in prior which makes EBSNet flexible and can adapt to more mobile applications.
        In order to train EBSNet, we treat it as an agent in RL and reward it by another neural network, MEFNet, which is used for multi-exposure fusion.
        By jointly training EBSNet and MEFNet, the proposed exposure bracketing selection can perform favorably against state-of-the-art methods both quantitatively and qualitatively. 
        We also provide a new dataset to facilitate future studies of exposure bracketing selection.

{\small
\bibliographystyle{ieee}
\bibliography{1804_final}
}

\end{document}